\documentclass[default,iicol]{sn-jnl}

\usepackage[algo2e]{algorithm2e} 
\usepackage{lscape}

\jyear{2021}%

\theoremstyle{thmstyleone}%
\theoremstyle{thmstyletwo}%

\theoremstyle{thmstylethree}%

\begin{document}

\title[Article Title]{The Design and Implementation of a Broadly Applicable Algorithm for Optimizing Intra-Day Surgical Scheduling}

\author[1]{\fnm{Jin} \sur{Xie}}\email{jinnyxie@stanford.edu}
\equalcont{These authors contributed equally to this work.}

\author[2]{\fnm{Teng} \sur{Zhang}}\email{tengz@stanford.edu}
\equalcont{These authors contributed equally to this work.}

\author[2]{\fnm{Jose} \sur{Blanchet}}\email{jose.blanchet@stanford.edu}

\author[2]{\fnm{Peter} \sur{Glynn}}\email{glynn@stanford.edu}

\author[5]{\fnm{Matthew} \sur{Randolph}}\email{mrandolph@stanfordchildrens.org}

\author*[2,3,4]{\fnm{David} \sur{Scheinker}}\email{dscheink@stanford.edu}

\affil*[1]{\orgdiv{Institute for Computational and Mathematical Engineering}, \orgname{Stanford University}, \orgaddress{\street{Via Ortega}, \city{Stanford}, \postcode{94305}, \state{CA}, \country{USA}}}

\affil*[2]{\orgdiv{Department of Management Science and Engineering}, \orgname{Stanford University}, \orgaddress{\street{Via Ortega}, \city{Stanford}, \postcode{94305}, \state{CA}, \country{USA}}}

\affil*[3]{\orgdiv{Clinical Excellence Research Center}, \orgname{Stanford University School of Medicine}, \orgaddress{\street{Campus Drive}, \city{Stanford}, \postcode{94305}, \state{CA}, \country{USA}}}

\affil*[4]{\orgdiv{Department of Pediatrics}, \orgname{Stanford University School of Medicine}, \orgaddress{\street{Welch Rd}, \city{Palo Alto}, \postcode{94304}, \state{CA}, \country{USA}}}

\affil*[5]{\orgdiv{Lucile Packard Children's Hospital}, \orgname{Stanford University}, \orgaddress{\street{Welch Rd}, \city{Palo Alto}, \postcode{94304}, \state{CA}, \country{USA}}}


\abstract{Surgical scheduling optimization is an active area of research. However, few algorithms to optimize surgical scheduling are implemented and see sustained use. An algorithm is more likely to be implemented, if it allows for surgeon autonomy, i.e., requires only limited scheduling centralization, and functions in the limited technical infrastructure of widely used electronic medical records (EMRs). In order for an algorithm to see sustained use it must be compatible with changes to hospital capacity, patient volumes, and scheduling practices. To meet these objectives, we developed the BEDS (better elective day of surgery) algorithm, a greedy heuristic for smoothing unit-specific surgical admissions across days. We implemented BEDS in the EMR of a large pediatric academic medical center.

The use of BEDS was associated with a reduction in the variability in the number of admissions. BEDS is freely available as a dashboard in Tableau, a commercial software used by numerous hospitals. BEDS is readily implementable with the limited tools available to most hospitals, does not require reductions to surgeon autonomy or centralized scheduling, and is compatible with changes to hospital capacity or patient volumes. We present a general algorithmic framework from which BEDS is derived based on a particular choice of objective and constraints. We argue that algorithms generated by this framework retain many of the desirable characteristics of BEDS while being compatible with a wide range of objectives and constraints. 
}




\maketitle

\section{Introduction}
\label{sec:intro}

Aggregated annual expenditures on surgical services in the United States, those including the direct cost of surgery and post-surgical hospitalization, are estimated at close to \$1 trillion. This figure amounts to about a quarter of the total health care costs in the U.S., which in turn are nearly 50\% higher on a per capita basis than those of any other country \cite{10.1001/jama.2018.1150}. The surgical environment is often the most expensive and highly resourced in the hospital, with one estimate putting the sum of direct costs, wages and benefits, and surgical supplies of one minute of operating room time at \$36 \cite{childers2018understanding}. Delays in access to recovery beds in the post anesthesia care unit, the intensive care unit, or the acute care unit are associated with adverse patient outcomes, surgical cancellations and delays, and higher spending. Improvements to the efficiency of surgical services and post-surgical recovery have the potential to meaningfully impact United States healthcare quality and expenditures.

At most US hospitals, the majority of surgical procedures are electively scheduled and offer an opportunity to smooth demand through better scheduling (although procedure-urgency mix varies across hospitals). For most US hospitals, surgical scheduling is a largely decentralized process in which each surgical service (e.g. neurosurgery or orthopedic surgery) is assigned  “blocks” of operating room time (e.g., two rooms 7am-5pm every Monday and Wednesday). Each block is then assigned to each surgeon in each service, e.g., a particular surgeon may have a room every Monday and Wednesday, or are used for the entire service, e.g., a Wednesday room may be reserved for urgent orthopedics procedures to be performed by whichever surgeon is available. Elective procedures are scheduled when a surgeon sees a patient in clinic and determines whether surgery is necessary and how soon it should occur. The clinic visit may be weeks to months in advance of surgery. In most hospitals, only a minority of procedures are emergent, ideally initiated within an hour of a patient’s arrival, or urgent, ideally performed within 24-48 hours of a determination of the need for surgery.    

Hospital managers and schedulers face four common challenges to optimizing operating room and postoperative bed utilization. The first is variation in scheduling lead times. Patients requiring urgent or emergent procedures arrive randomly and patients requiring elective procedures have different clinical needs and preferences for when to have the procedure. In the United States, where elective surgery is relatively profitable, hospitals face financial pressure to accommodate patient preferences. The second challenge is that surgical services and individual surgeons enjoy significant autonomy, limiting the decision making power of any central scheduler and creating challenges for implementing centralized optimized scheduling. Typically, each surgical service, e.g,. orthopedics, is allocated operating room time in blocks that are then assigned to individual surgeons within the service. Each surgeon enjoys significant discretion in how to use their blocks, which allows them to prioritize patients based on clinical acuity, accommodate patient preference, and attend to other professional commitments. However, this flexibility may also create perverse incentives that may lead to inefficient use of OR time. For example, if a surgeon plans to attend a conference, she may make informal arrangements to swap blocks with other surgeons or, if she does not bare the costs of low OR utilization, may leave her block unused on the off chance she does not attend the conference. The third, closely related, challenge is that surgeons and operating rooms are non-fungible and subject to frequent changes in availability. A large institution may have to deal with the resignation or hiring of one or more surgeons per month and at any given time one or more operating rooms may be closed for repairs or upgrades. New surgeons bring their own patient mix and may require access to different resources and operating room blocks. Approaches to scheduling optimization that incorporate surgeon-specific profiles (e.g., average patient post-operative length of stay) and operating room characteristics (e.g., only some operating rooms are suitable for cardiac surgery) may require relatively frequent updating corresponding to such changes. The fourth challenge is that most hospitals have limited technical infrastructure and expertise to implement or update sophisticated scheduling algorithms. EPIC and Cerner, the two largest US electronic medical record (EMR) vendors, account for over 65\% of patient beds in the US but offer tools to operationalize only a few simple machine learning algorithms, no functionality to implement optimization algorithms such as dynamic programming, and limited or no functionality to automate the import of optimized schedules generated externally. Efforts to optimize scheduling often rely on extracting data from the EMR, processing it externally to generate optimal schedules, and a manual process to modify the schedule to match the one produced through optimization.

Two principled approaches to optimizing operating room and postoperative bed utilization are to change the block schedule or to change how surgeries are scheduled into a fixed block schedule. Each presents challenges associated with complex constraints and changes to surgical demand and hospital resources, and relatively few have been implemented \cite{samudra2016scheduling}.

The surgical block schedule coordinates the availability of numerous stakeholders and resources and, in some institutions, may remain largely unchanged for decades. There have been a few successful examples of the use of operations methodology, simulation and integer programming, to redesign the block schedule of a major hospital to reduce wait times for non-elective cases and smooth the demand for postoperative beds; see \cite{zenteno2015pooled, zenteno2016systematic}. These projects required detailed models representing the hospitals at which the work took place. They utilized analytical resources beyond those available to the hospital. These studies document a number of significant practical barriers involved in changing block schedules and the sustained leadership required to make the appropriate changes in block scheduling; including altering surgeon research and clinical schedules. Non-stationarity in the surgical volumes add to the practical considerations discussed in \cite{zenteno2015pooled, zenteno2016systematic} with regard to the time required to implement changes in blocks. Both works describe the difficulty of evaluating the impact of the interventions presented as a result of significant variations in surgical volume and changes in surgical case mix. The policy described in \cite{zenteno2016systematic} is updated every couple of years to account for the changes in volume as well as the availability of surgeons, operating rooms, and patient characteristics.

Coordinating surgical scheduling into a fixed block schedule faces similarly complex challenges. Surgeons usually enjoy significant independence in scheduling cases into their assigned blocks. This allows surgeons to prioritize patients based on clinical urgency and difficulty. A surgeon may decide that for some elective cases it is safe to wait months while others should be completed in days; a surgeon may prefer performing difficult cases early in the morning; or a patient may require concurrent procedures from more than one surgeon requiring that surgeon schedules be coordinated. A few operations-based approaches to surgical scheduling, using machine learning, optimization, and simulation, have been developed and deployed \cite{fairley2019improving}. Significant effort is required to generate detailed models that capture the idiosyncratic constraints of the institutions where the work is performed, develop specialized software to deploy the scheduling algorithms, and secure the cooperation of surgeons whose incentives may not be fully aligned with the overall efficiency of the system \cite{fairley2019improving}. The work in \cite{scheinker2020implementing} documents how the combination of complex constraints and non-stationary system characteristics led to the suspension of the use of the algorithm designed in \cite{fairley2019improving}. In particular, the institution opened new operating rooms, opened new recovery beds, and repurposed existing operating rooms and recovery beds far faster than the tool could be updated and redeployed.

Institutions seeking to improve surgical block allocation or optimize scheduling have to develop models that incorporate their operational constraints, develop internal analytics expertise or partner with an external vendor, and overcome political and change-management barriers to deployment. Furthermore, they must be able to do so sufficiently quickly to update or redesign the models as their volumes or resources change. This may explain why, among the few papers reporting an optimization model implemented at a single institution, we found no publications documenting the successful reproduction of the work at other institutions. 

\subsection{Contribution}

We develop the BEDS (Better Elective Day of Surgery) algorithm, a simple heuristic for inter-day surgical scheduling. We show that it has the following desirable properties: 
\begin{enumerate}
    \item Compatible with a common surgical scheduling workflow
    \item Requires as input a minimal set of routinely collected data
    \item Produces interpretable recommendations
    \item Does not require significant reductions to surgeon autonomy or centralized scheduling
    \item Is compatible with changes to demand, capacity, and operations such as surgeon vacations, illness, or joining/leaving the hospital workforce; long-term changes in surgical volumes; the opening or closing of new operating rooms and post-procedural beds; and the introduction of intra-day scheduling optimization
    \item The implementation of BEDS at an academic medical center was associated with reduced variability in the daily number of post-surgical admissions
    \item It is readily implementable in the limited technical capabilities of modern EMRs
    \item A free version is available for use on Tableau Server, a commercial software widely used by hospitals in the US.
\end{enumerate}

We also present the BEDS framework, of which the BEDS algorithm is one application. We think of the BEDS framework as a structured formalism that can be use to guide the design of algorithms that retain many of the desirable characteristics of BEDS while being compatible with a wide range of objectives and constraints. Algorithms derived from the BEDS framework are readily implementable for surgical scheduling and may serve as benchmarks for more sophisticated scheduling algorithms.



The rest of the paper is organized as follows. We provide a review of the literature on surgical scheduling in Section \ref{Section_Literature}. We discuss the technical challenges involved in surgical scheduling under non-stationary settings in Section \ref{sec:challenges}. We describe the BEDS algorithm in detail in Section \ref{Section_BEDS_Algo}. In order to test and validate the BEDS algorithm and inform our hospital's decision to potentially adopt BEDS, we constructed a discrete event simulation model based on historical data, which is described in Section \ref{sec:methods}. The validation results of the simulation model are described in Section \ref{sec:results}. The results of implementation at the hospital are discussed in Section \ref{sec:implementation}. We present the general BEDS framework in \ref{Section_BEDS_Frame}. Final considerations and conclusions are given in Section \ref{sec:discussion}. Technical details such as data processing and data structure architectures for our validation, testing and simulations are provided in the appendix Section \ref{sec:appendix}.

\section{Literature Review} \label{Section_Literature}

\noindent
Surgical scheduling has been an active research area for decades \cite{guerriero2011operational, zhu2019operating, rahimi2020comprehensive}. Traditionally, the surgical operating room (OR) scheduling problem has been studied with operating room time as the single resource to be allocated and a single objective such as minimizing overtime, maximizing operating room utilization, and minimizing patient wait time, etc. However, surgical patients do not interact with the hospital only through the OR. Patients require many resources including post-operative units such as the post anaesthesia care unit (PACU), intensive care unit (ICU), and acute care unit or ward. Surgical planning with consideration of downstream unit demand has attracted significant attention in recent years. 

Two lines of research relate to our work in designing surgical scheduling that consider post-procedural units. One stream of work relies on the idea of directly adding the post-procedural unit capacity as constraints of the mathematical problem for OR-centered optimization, see \cite{shehadeh2020stochastic} for a comprehensive review of recent developments along this line of research. Another type of work adopts the idea of using the performance metrics of downstream units as the objective function of the surgical scheduling optimization. Our work falls into the second category, for which we provider a closer review of the literature.

\begin{sidewaystable*}
\sidewaystablefn%
\begin{center}
\begin{minipage}{\textheight}
\small{
\begin{tabular}{lllllll}
\hline
\multicolumn{1}{c}{Setting}                                                            & \multicolumn{1}{c}{Publication} & \multicolumn{1}{c}{\begin{tabular}[c]{@{}c@{}}Post-Op \\ Unit\end{tabular}} & \multicolumn{1}{c}{Objective}                                                                                                                                           & \multicolumn{1}{c}{\begin{tabular}[c]{@{}c@{}}Modeling \\ Method\end{tabular}} & \multicolumn{1}{c}{\begin{tabular}[c]{@{}c@{}} Non-stationary\\ Inputs\end{tabular}}                                                                                                                                        & \multicolumn{1}{c}{\begin{tabular}[c]{@{}c@{}}Reported\\ Implementation\end{tabular}} \\ \hline
\multirow{5}{*}{\begin{tabular}[c]{@{}l@{}}master \\ surgery \\ schedule\end{tabular}} &                       \cite{belien2007building}          & Ward                                                                        & min (census variance)                                                                                                                                                   & MIP                                                                            & \begin{tabular}[c]{@{}l@{}} expected patient volume\end{tabular}                                                                                          &                                                                                    \\
                                                                                   &      \cite{price2011reducing}                           & ICU                                                                         & min (expected admission - discharge)                                                                                                                                    & MIP                                                                            & \begin{tabular}[c]{@{}l@{}}service line grouping, \\ expected patient volume\end{tabular}                                                                  &                                                                                    \\
                                                                                       &                                \cite{chow2011reducing} & Ward                                                                        & min (max bed census)                                                                                                                                                    & MIP                                                                            & \begin{tabular}[c]{@{}l@{}}expected patient volume\end{tabular}                                                                                              & Y                                                                                  \\
                                                                                       &  \cite{zenteno2016systematic}                               & Ward                                                                        & max (reduced peak bed census)                                                                                                                                           & MIP                                                                            & \begin{tabular}[c]{@{}l@{}} expected patient volume\end{tabular}                                                                                              & Y                                                                                  \\
                                                                                       &  \cite{belien2009decision}                               & Ward                                                                        & \begin{tabular}[c]{@{}l@{}}multiple weighted objectives: \\ min (peak bed utilization)\\ min (bed utilization variance)\end{tabular}                                    & MIP                                                                            & \begin{tabular}[c]{@{}l@{}} expected patient volume\end{tabular}                                                                                              & Y                                                                                  \\ \hline
\multirow{4}{*}{\begin{tabular}[c]{@{}l@{}}advance \\ schedule\end{tabular}}           &  \cite{bekker2011scheduling}                               & Ward                                                                        & min ($\vert$scheduled load - target load$\vert$)                                                                                                                                      & \begin{tabular}[c]{@{}l@{}}Queueing\\ MIP\end{tabular}                          & \begin{tabular}[c]{@{}l@{}} expected patient volume\end{tabular}                                                                                          &                                                                                    \\
                                                                                       &   \cite{van2020minimizing}                              & Ward                                                                        & \begin{tabular}[c]{@{}l@{}}multiple weighted objectives: \\ min (bed utilization variance), \\ min (wait list size), \\ max (number of schedule surgeries)\end{tabular} & MIP                                                                            & \begin{tabular}[c]{@{}l@{}}procedure distribution,  \\ patient wait list\end{tabular}                                                                     &                                                                                    \\
                                                                                       &   \cite{jebali2017chance}                              & ICU                                                                         & \begin{tabular}[c]{@{}l@{}}multiple weighted objectives: \\ min (patient-related cost)\\ min (OR overtime/undertime)\\ min (ICU over-capacity)\end{tabular}             & CCSP                                                                           & \begin{tabular}[c]{@{}l@{}}procedure  distribution, \\ patient cost function,\\ OR utilization cost function, \\ emergency cases volume\end{tabular} &                                                                                    \\
                                                                                                                                                            &   \cite{fairley2019improving}                              & PACU                                                                         & \begin{tabular}[c]{@{}l@{}}multiple weighted objectives: \\ min (finishing time) \\ min (PACU occupancy)\end{tabular}             & IP                                                                           & 
                 \begin{tabular}[c]{@{}l@{}} OR room availability \\ post-op beds availability \end{tabular}                 &                                                                                    \\
                                                                                       & \textbf{BEDS}                   & \textbf{\begin{tabular}[c]{@{}l@{}}ICU, \\ Ward\end{tabular}}               & \textbf{min (admission variance)}                                                                                                                                       & \textbf{heuristic}                                                             & \textbf{\begin{tabular}[c]{@{}l@{}}
                                                                           \end{tabular}}                                                                              & \textbf{Y}                                                                         \\ \hline
\end{tabular}
}
\caption{Comparison of related papers. By ``expected'', we mean a point forecast for the mean.}
\label{table:paper-comparison}
 \end{minipage}
 \end{center}
 \end{sidewaystable*}

We summarize the related literature together with our work in Table~\ref{table:paper-comparison} as a comparison. All papers propose a solution to the problem of minimizing variability in the post-op units as the objective by optimizing surgical schedules, and are based on the surgical block system. 
We categorize by two different settings: master surgery schedule (MaSS), in which the assignment of blocks to services is optimized, and advance schedule (AdS), in which scheduling procedures into fixed blocks is optimized. For each publication we show its post-op unit considered, the modeling objective and method, information needed to implement the proposed method and whether the solution had been implemented in a hospital at the time of publication. 

In the master surgery schedule (MaSS) literature, \cite{belien2007building} proposes using length of stay (LOS) distributions to calculate the distribution of the patient census given patient volume, then directly minimize the variance within a mixed integer program. Based on this idea, follow up work instead uses only certain statistics, such as the expected census, other than the entire distribution. Objectives such as the minimization of the peak census are also considered in \cite{price2011reducing, chow2011reducing, zenteno2016systematic, belien2009decision}. 

In the AdS literature, \cite{bekker2011scheduling} provides a queuing theory-based method of estimating the optimal daily number of admissions accommodating a lower number of staffed beds during the weekends, and then solving a mixed integer programming (MIP) problem with the objective minimizing the difference between scheduled load and target load. A key assumption underlying this queuing theory analysis is stationarity, which does not necessarily hold in many applications. Other work, including \cite{van2020minimizing, jebali2017chance}, consider the stochasticity of LOS and surgical duration, and add the post-op units performance metric into the weighted objective function. Via different choices of institutional objectives and constraints, \cite{van2020minimizing} formulates the scheduling problem as a two-stage MIP and \cite{jebali2017chance} formulates the problem as a chance-constrained stochastic program. Our method, compared to 
these methods, is simple enough that we do not need to solve any mathematical programming problem and thus easy to implement and incorporate into the current system that hospitals are deploying.

\section{Scheduling Challenges in Highly Connected and Non-stationary Environments}
\label{sec:challenges}

In order to motivate the scheduling heuristic that we introduce in the next section, we describe in this section the challenges that arise in finding a desirable scheduling policy from a practical perspective. 

In general, the resource management of an operating room (OR) is a decentralized yet highly inter-connected process. 
As a result, most existing works propose a centralized approach, which is usually realized by a large mathematical programming problem that incorporates all surgeon/services and independent stakeholders in the system (as in Table~\ref{table:paper-comparison}). 
When modeling a system with multiple resources and stakeholders, like the hospital or other medical institutions, researchers are inclined to integrate as much information as possible so that the optimal solution produced by the optimization model can achieve comparable performance as in real-world systems. 
As we can see in Table~\ref{table:paper-comparison}, to generate a good advance scheduling policy, complicated mathematical optimization problems are constructed with complex objective functions and integrated constraints that require significant implementation efforts. For example, \cite{zenteno2016systematic} reports the following challenge when implementing a MIP (Mixed Integer Programming)-based block scheduling optimizer at the
Massachusetts General Hospital (MGH). The authors found that some scheduled times produced by the integer programming program were actually infeasible in the real-world system due to constraints not initially incorporated by the model such as surgeon unavailability due to academic, administrative or clinical reasons. As a result, several rounds of iteration and negotiation with surgical leadership were required as part of 6-month effort to incorporate additional constraints into the model or change surgeon schedules.  

Hospitals routinely experience changes in their surgical workforce, changes to the physical operating room space and changes to patient volume and type. In \cite{scheinker2020implementing, zenteno2015pooled, zenteno2016systematic}, such changes, including changes in volume and the opening of new operating rooms, are described as part of the reason why it was difficult to evaluate the impact of the interventions. At the same time, in mathematical optimization, especially integer programming models (which are the models adopted by most works in Table~\ref{table:paper-comparison}), when the objective or a constraint changes, the original solution may no longer be optimal or feasible \cite{zenteno2016systematic}. Maintaining such models requires frequent updates when, for example, a new surgeon is hired, a new building/unit is opened, the number of operating or recovery rooms changes, a surgeon's schedule changes, a surgeon's block or clinic schedule changes, etc. If updating the algorithm and the tool through which it is deployed requires significant effort, it may fall out of use as was the case in \cite{fairley2019improving}.

As reported in \cite{zenteno2016systematic}, a more serious problem arises when there is no feasible solution of the mathematical program after new constraints are introduced. In this case, the researchers have to make significant efforts based on institutional-specific needs to deal with the trade-off between optimizing the institutional goal and mathematical soundness of the algorithm. Compared to robust optimization methods applied in surgical scheduling problems as in \cite{two-stage}, the approach we propose in the next section is easier to implement at the hospital, since it does not require a fixed time horizon for an offline optimization problem or batch scheduling of the the patients in implementation. The method we propose is an online heuristic that is applied each time a patient is scheduled. Meanwhile, as pointed out in \cite{two-stage}, robust optimization methods can be over-conservative in its application to surgical scheduling problems, which might lead to significant surgical delays in practice.




















\section{BEDS Algorithm}\label{Section_BEDS_Algo}

We propose the BEDS (Better Elective Day of Surgery) algorithm \ref{alg: BEDS}. At the institution studied, the objective was to minimize the maximum total number of daily admissions while minimizing patient wait time for surgery. We designed BEDS to embed in a common surgical scheduling workflow. A surgeon examines a patient in clinic, recommends a type of surgery, and estimates how soon it should be performed. The patient then works with a scheduler to choose a day of surgery. BEDS performs a greedy-search among the feasible surgical days to minimize current post-op admissions and patient wait time. The only data required are: clinically acceptable wait time for the procedure, surgeon availability, patient availability, estimated procedure duration, and occupancy data for the post-operative unit. 

Only a small set of input data are required for implementation (Tables \ref{data: BEDs1}, \ref{data: BEDs2}, and \ref{data: BEDs3}).

Note that the greedy recommendation step of the algorithm is modular; the logic may be modified without perturbing the remainder of the algorithm. In the general BEDS framework the recommendation is modifiable based on the objectives or constraints of the institution or surgeon where it is to be deployed. For example, how much patient preferences shape the process may be modified by the institution by having the algorithm return the $n$ top-ranked days to be offered to the patient to choose from.

Many US hospitals use tableau (While specific numbers were not available, Tableau advertises that all of the top 20 US News \& World Report ranked hospitals do). We have created a tableau workbook, available on Github (\url{https://github.com/tengz-sudo/BEDS\_Tableau}), for other hospitals to deploy BEDS on their tableau server. Below, as shown in Figure 1, is an example of the user interface when the recommendation is set to present all feasible days color-coded by the number of admissions. When a patient comes to the clinic, the scheduler uses BEDS to choose the patient's primary surgeon, target post-procedure unit, and to filter out days on which the surgeon lack sufficient time for the case. The different colors on the heat map represent the number of patients already scheduled to be admitted to the target unit on the same day (green - fewest patients). The scheduler can then work with the patient to find a suitable ``green" day. A different choice of objective and ranking would require only a modification to the definition of the color coding.

\begin{algorithm}[H]
\caption{BEDS}
\label{algorithm:BEDS}
 \SetAlgoLined
    For each day $d$ in the scheduling horizon $D$, require: 
    \begin{enumerate}
        \item For each surgeon $j$, their number of available hours $h_{d,j}$.
        \item For each post-op unit $u$, the number of patients $n_{d,u}$ scheduled to be admitted.
    \end{enumerate}
    
\For{patient $i$ being scheduled for a procedure}{
    Determine that:
    \\
    \text{\quad}1. Surgeon $j$ that will perform the procedure.\\
    \text{\quad}2. Scheduled procedure duration, $g_{i}$.\\
    \text{\quad}3. Clinical and institutional window of \\
    \text{\quad}  availability for the procedure, $[L_i, R_i]$.\\
    \text{\quad}4. Patient's window of availability for the \\
    \text{\quad}procedure, $[l_i, r_i]$.\\
    \text{\quad}5. The unit $u_i$ to which the patient will \\
    \text{\quad}be admitted after the procedure.
    \\
    Based on the above:\\
    \text{\quad}1. The candidate dates $\mathcal{D}_i$ are those within\\
    \text{\quad}the windows of availability on which the \\
    \text{\quad}surgeon has sufficient time, i.e., $h_{d,j} \ge g_{i}$.
    \text{\quad}2. \textbf{Greedy recommendation} Of the days \\
    \text{\quad} $d\in\mathcal{D}_i$ that achieves the minimal $n_{d,u}$, \\
    \text{\quad}return the earliest. \\
    \text{\quad}3. Update $h_{d_{i},j}$, $n_{d_{i},u}$, and $h_d$.
}
\label{alg: BEDS}
\end{algorithm}

\begin{table}[h]
\centering
\begin{tabular}{|l|l|}
\hline
Column  & Value \\ \hline
Date & 2015-02-08\\ \hline
Surgeon ID & 1\\ \hline
Available Hours & 5.5 \\ \hline
\end{tabular}
\caption{The header of the \textit{surgeon hour} data.}
\label{data: BEDs1}
\end{table}

\begin{table}[h]
\centering
\begin{tabular}{|l|l|}
\hline
Column  & Value \\ \hline
Date & 2015-02-08\\ \hline
Unit & PCUs\\ \hline
Number of Scheduled Patients& 3 \\ \hline
\end{tabular}
\caption{The header of the \textit{unit occupancy} data.}
\label{data: BEDs2}
\end{table}

\begin{table}[h]
\centering
\begin{tabular}{|l|l|}
\hline
Column  & Value \\ \hline
Primary CSN & 1\\ \hline
Primary Surgeon ID & 1\\ \hline
Available Window & [2015-02-01, 2015-02-07]\\ \hline
Surgery Duration& 2.5 \\ \hline
Post-op Unit& PCUs \\ \hline
\end{tabular}
\caption{The header of the \textit{patient availability} data. The column 'Primary CSN' is a unique ID for each patient visit to the hospital.}
\label{data: BEDs3}
\end{table}

\begin{figure*}[h]
    \begin{center}
     \includegraphics[width = 14cm]{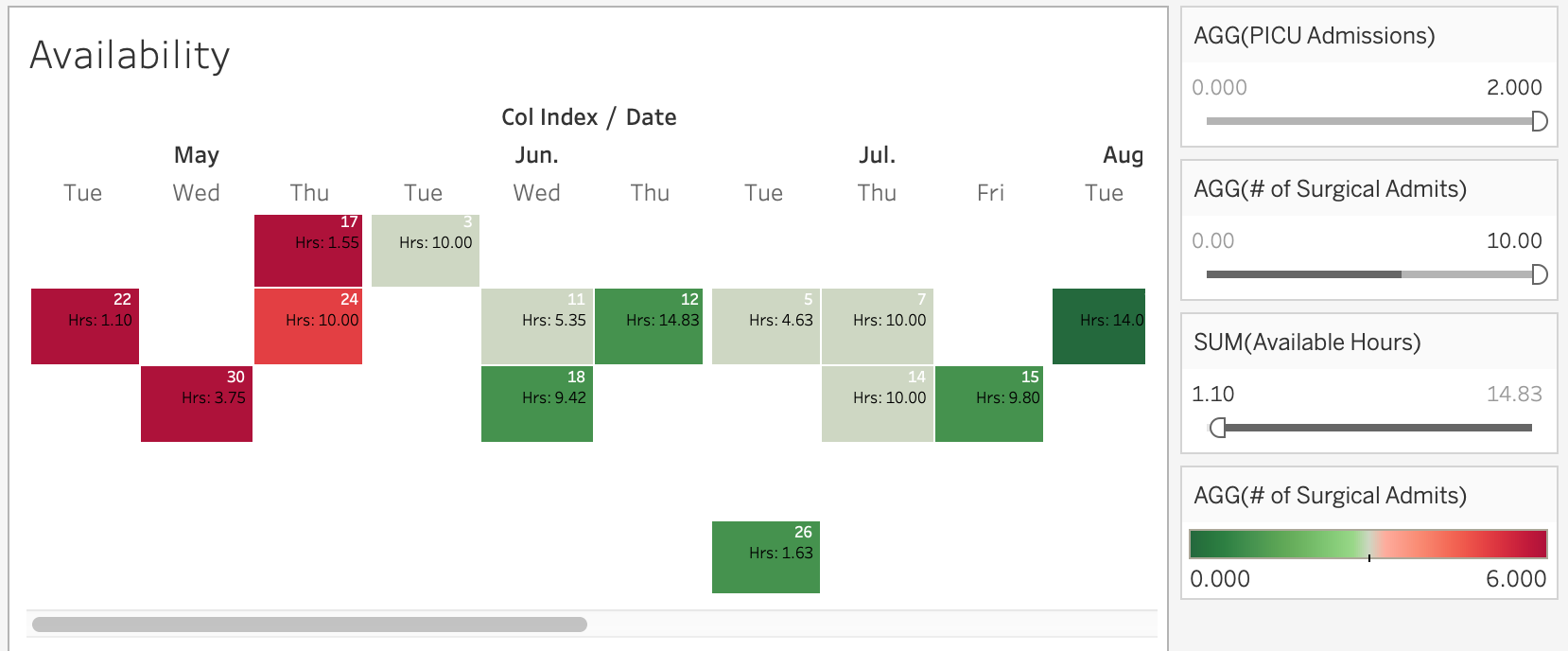}
     \caption{In this figure, we show a calendar heatmap in Tableau, which is our user interface of the scheduling tool. The heat map is a simple, color coded calendar. Using the heatmap as a guide, schedulers will select a few days that have a low number of surgery admissions to propose to the patient. The schedulers will then contact the patient to confirm a date of surgery.  The main function of BEDS is the heatmap to visualize the best and worst days to schedule patients for surgery based on the number of currently scheduled surgical patients with the same in-hospital postoperative recovery destination. The color of the block of a specific day represents the number of surgical admits already scheduled for that day. The color code of the number of surgical admits is at the bottom right. In each block of a specific day, the date is marked on the block in white and the number of remaining hours a surgeon has for that day is marked in black.
}       
    \end{center}
    \label{fig: BEDs}
\end{figure*}

\begin{figure*}[h]
    \begin{center}
    \includegraphics[width = 15 cm]{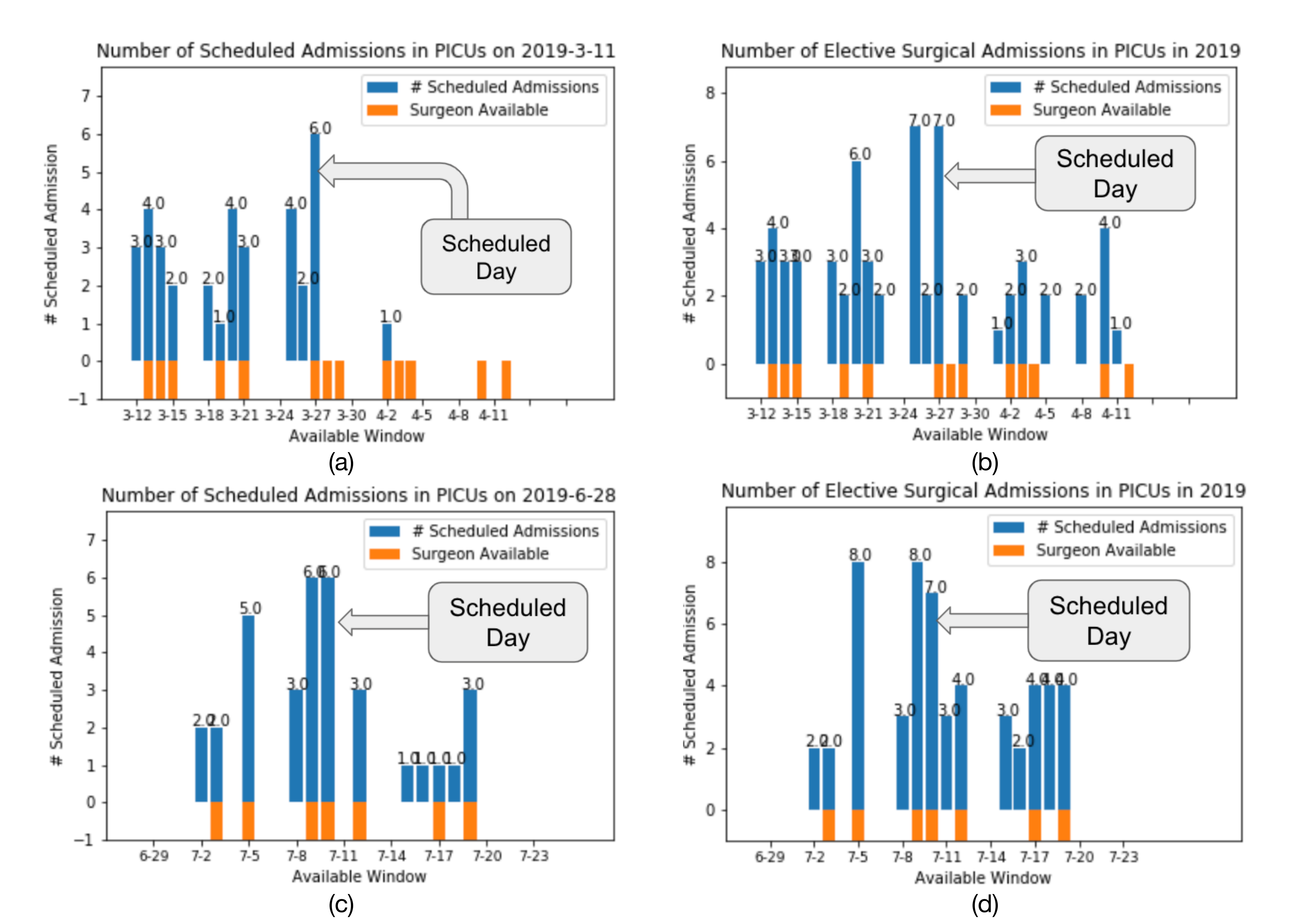}
    \caption{In this figure, we show two example patients for whom the BEDS scheduling heuristic can be helpful for scheduling his/her admission to a day with less scheduled admissions in the patient's available window. Each patient was scheduled for a date on which six surgical admissions had already been scheduled for the PICU, despite the surgeon having availability on days with significantly fewer scheduled cases (as in (a) and (c)). The first patient requested a schedule for his admission to the hospital on 2019-03-11 and was scheduled to a day (2019-03-27) on which six surgical patients have already been scheduled to be admitted. The second patient requested a schedule for his admission to the hospital on 2019-06-28 and was scheduled to a day (2019-07-10) on which six surgical admissions have already been scheduled. For the both of them, there are days within their available window, as the days marked on the x-axis, when their surgeons are available, as the orange bar indicates, and there are less than six surgical admissions scheduled. Predictably, on the days the patients were admitted, the total number of admissions to the pediatric ICU was significantly higher than on other days when the surgeon was available (as in (b) and (d)). }       
    \end{center}
    \label{fig: example}
\end{figure*}

In the use case at our institution, BEDS was designed to help schedulers avoid inadvertently selecting a day for which many patients requiring admission to the same unit have already been scheduled. As an example, we show real historical data for patients scheduled in March and June of 2019 for procedures after which they required admission to the pediatric intensive care unit (PICU) (Table \ref{table: example}). Each was scheduled for a date for which six surgical admissions had already been scheduled for the PICU, despite the surgeon having availability on days with significantly fewer scheduled cases (Figure \ref{fig: example}  left). Predictably, on the days the patients were admitted the total number of admissions to the pediatric ICU was significantly higher than on other days when the surgeon was available (Figure \ref{fig: example}  left). 
\begin{table*}[h]
\centering
\begin{tabular}{|l|l|l|l|}
\hline
Primary CSN  & Surgery Request Date & Surgery Date & Assumed Available Window of Patient\\ \hline
1  & 2019-03-11 & 2019-03-27 & [2019-03-12, 2019-04-12]\\ \hline
2  & 2019-06-28 & 2019-07-10 & [2019-06-29, 2019-07-22]\\ \hline
\end{tabular}
\caption{Examples for the Scheduling Heuristic.}
\label{table: example}
\end{table*}


\section{Simulation Model}
\label{sec:methods}
In order to build an overall evaluation tool to analyze the impact of different scheduling policies on the patient flow, we build a discrete event simulation (DES) model. The simulation model allows us to combine different data resources like the midnight census data, procedure data, and staff availability data, etc. It generates a discrete event queue representing the patient flow under specific scheduling logistic. Each event in the queue is a representation of the admission, transfer between units, or discharge of a patient who visits the hospital (summarized in Table~\ref{table:events}). For the scope of units in the simulation, our model focuses on pediatric areas (acute care and ICU areas). In this section, we elaborate on the high-level design and mechanism of our discrete event simulation model (summerized in Figure 2) with more details given in Section \ref{sec: simulation model more}.

To use the simulation model, it is required that the user provides the \textit{midnight census} data as well as the \textit{procedure record} data of their medical institution as inputs. The \textit{midnight census} data are records of all the patients staying in the hospital overnight with an assigned bed at midnight. The \textit{procedure record} data are records of every surgery that happen at the hospital (This includes the time, duration, type, who performed the surgery and in which operating room), see Section \ref{app_sim}. The user has an option of either providing \textit{surgeon availability} data or choosing a built-in function that generates \textit{surgeon availability} from the data \textit{procedure record}. We summarize the inputs of the model in Table \ref{data: BEDs1}, \ref{data: BEDs2} and \ref{data: BEDs3}.

\begin{table}[h]
\centering
\begin{tabular}{|l|l|}
\hline
Data Inputs & Required \\ \hline
\textit{midnight census} & Yes \\ \hline
\textit{procedure record}& Yes \\ \hline
\textit{surgeon availability} & No\\ \hline
\end{tabular}
\caption{Data inputs for the simulation model.}
\label{table: inputs}
\end{table}

\begin{table*}[h]
\centering
\begin{tabular}{lll}
\hline
\multicolumn{1}{c}{Event Name} & \multicolumn{1}{c}{Meaning}                                                                             & \multicolumn{1}{c}{Next Event}                                                     \\ \hline
ARRIVAL                        & \begin{tabular}[c]{@{}l@{}}elective surgical patient request schedule\\ /outpatient visit\end{tabular} & TRANSFER\_IN                                                                       \\
TRANSFER\_IN                   & patient transfer into some unit                                                                         & \begin{tabular}[c]{@{}l@{}}READY\_TO\_TRANSFER\end{tabular} \\
READY\_TO\_TRANSFER            & patient ready to transfer out from some unit                                                            & TRANSFER\_IN                                                                       \\
DISCHARGE                      & patient discharge from the hospital                                                                     &                                                                                    \\ \hline
\end{tabular}
\caption{Events considered in the simulation. The simulator generates the next event based on the current event type and follows historical data or a random time generator bounded in given time interval.}
\label{table:events}
\end{table*}

\begin{figure*}[h]
    \begin{center}
     \includegraphics[width = 12.5 cm]{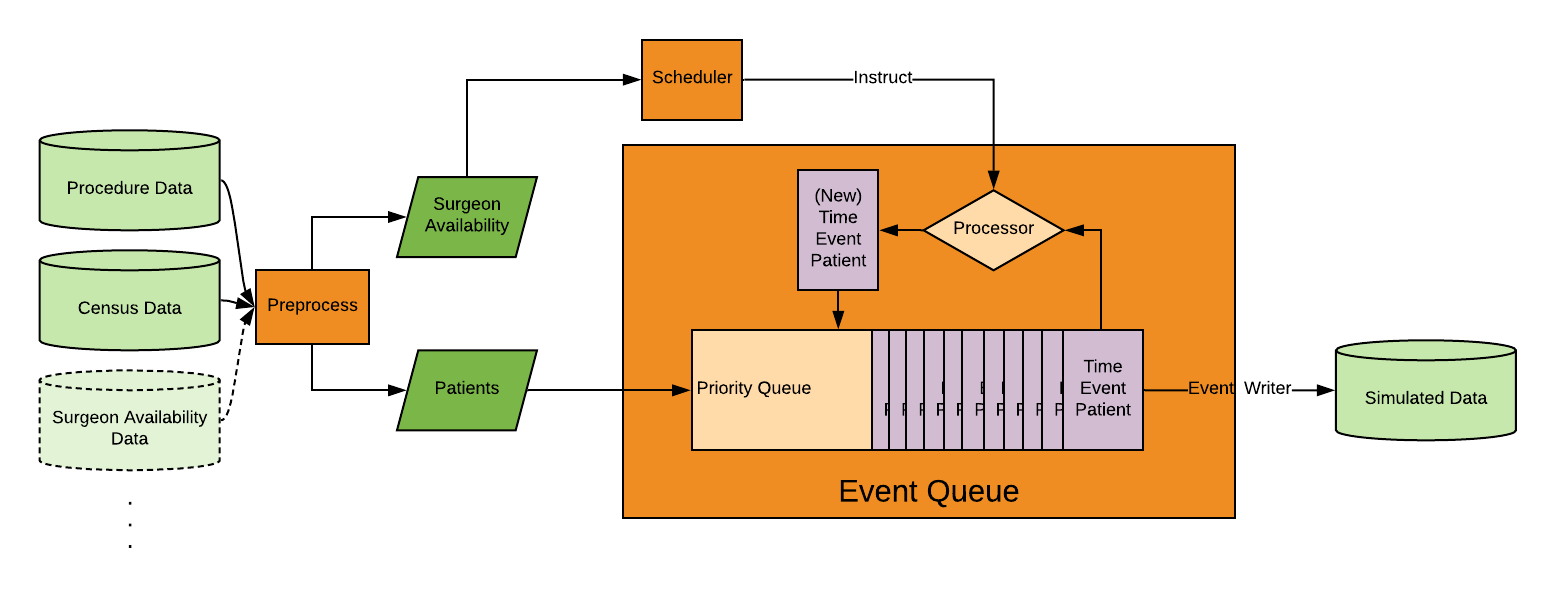}
    \caption{In this figure, we show the discrete event simulation (DES) model outline. As in the three modules on the left, we first collect the \textit{procedure} data, the \textit{census} data, and the \textit{surgeon availability} data (if applicable) and feed all of them into the data pre-processor. Then the data pre-processor combines all the different data resources and reorganize them to two data frames: the \textit{patients} data and the \textit{surgeon availability} data. Then, the data frames are fed to a scheduler module and a priority queue is generated by a processor module as in the box in the middle. In the end, an event writer records the information of the event queue in the simulated data as the output of the model.}       
    \end{center}
    \label{fig: model_struct}
\end{figure*}

The major steps of simulation is summarized as follows:
\begin{itemize}
    \item \textbf{Step 1}: Feed the input data \textit{midnight census}, \textit{procedure record} and \textit{surgeon availability} (if applicable) to the pre-processor and generate the data \textit{patient profile} and \textit{surgeon availability} (if not provided). Each row of the data \textit{patient profile} summarizes the trajectory of one patient visit in the hospital, see Table~\ref{table:patient-profile-header} in Section \ref{app_sim}.
    \item \textbf{Step 2}: Use the \textit{patient profile} and \textit{surgeon availability} data to generate the discrete event queue under the scheduling logic encoded in the Scheduler.
    \item \textbf{Step 3}: Output the simulated event series and generate the simulated hospital record data for further data analysis.
\end{itemize}

\section{Simulation Results}
\label{sec:results}
In this section, we present the results of simulating the patient flow at the Lucile Packard Stanford Children's Hospital (LPCH) from January 2019 to March 2021 using the simulation model described in Section \ref{sec:methods}. We model the outcome of implementing BEDS at the hospital with two experiments, as described in Table \ref{table: experiments}. In both experiments, we apply BEDS to schedule all elective surgical patients to be admitted to the pediatric intensive care units (PICUs) or acute care units (PCUs).
We include all the patient records with arrival times from 2018-01-01 to 2018-12-31 as a one-year warm-up period for the simulation system since the occupancy of patients admitted during the warm-up period can affect the scheduling of the patients admitted afterwards. We include the warm-up period  in the simulation system in order to obtain a valid initialization of the system. Otherwise, those patients who are already scheduled for surgery in the real hospital system would not be captured in the simulation system. For data pre-processing, we exclude patients and records with important missing information as described in Section \ref{app_sim}. As a result, $99.8 \%$ of the total 334274 records in the \textit{midnight census} data are included and none of the total 5558 surgical patients to be admitted to PICUs or PCUs in the record during the period of simulation is excluded in data pre-processing.

In the first experiment, the scheduler of the simulation model is set to the `BEDS' mode on 2020-07-27, as in the real world implementation. Before that, the scheduler assigns patients to their original scheduled time in the hospital record. This experiment tests whether our simulation model reproduces historical hospital data and models the outcome of scheduling policies.
 
\begin{figure*}[h]
    \centering
    \includegraphics[width = 11cm, height=8cm]{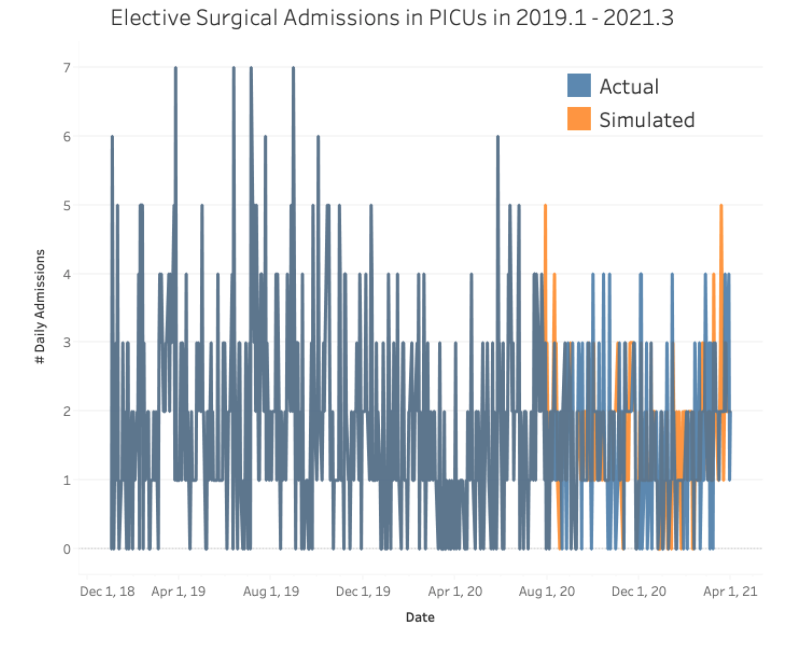}
    \caption{Comparison of the number of daily elective surgical patient admissions into PICUs in actual and simulated data in 2019.1 - 2021.3. The actual historical data is as the blue line and the simulated data is as the orange line. Before the BEDS implementation on 2020.7.27, the simulated patient flow reproduces the actual historical data.}
    \label{fig:PICU21_0}
\end{figure*}

\begin{figure*}[h]
    \centering
    \includegraphics[width = 15 cm]{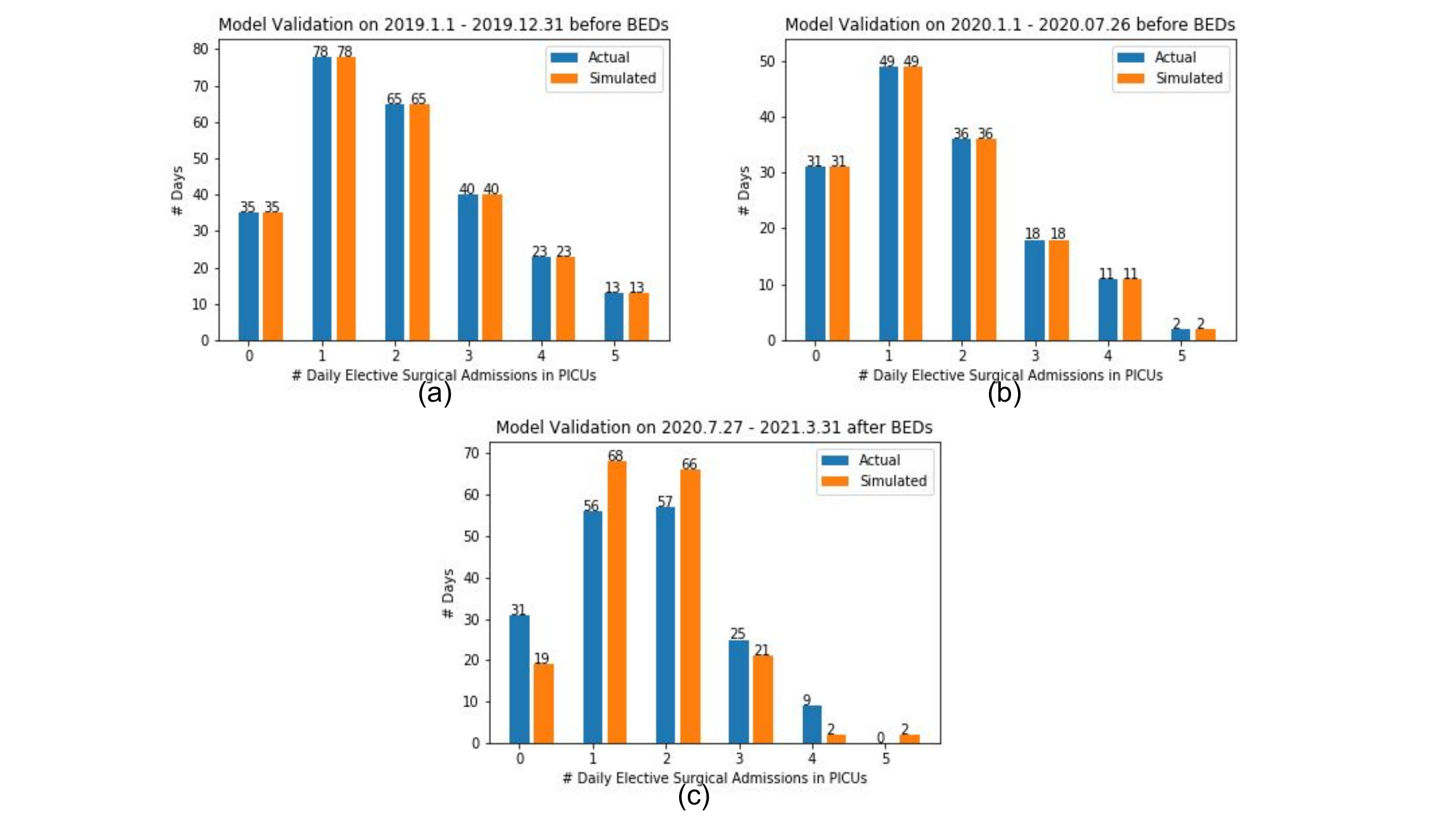}
    \caption{Comparison of the histogram of the number of daily elective surgical patient admissions into PICUs in historical and simulated data in 2019.1 - 2021.3. We show the histogram of three different periods: 2019.1.1 - 2019.12.31 (a), 2020.1.1 - 2020.7.26 (b) before BEDS implementation, and 2020.07.27 - 2021.3.31 (c) after BEDS implementation. The actual historical data is as the blue bars and the simulated data is as the orange bars.}
    \label{fig:PICU21}
\end{figure*}

\begin{figure*}[h]
    \centering
    \includegraphics[width = 7 cm]{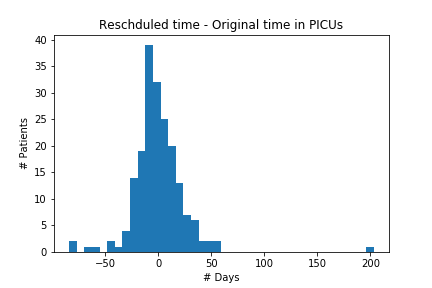}
    \caption{Difference (by days) between rescheduled time and original time in record for patients rescheduled in the simulation model in PICUs during 2020.7.27 - 2021.3.31 after BEDS implementation.}
    \label{fig: diff}
\end{figure*}

\begin{table}[h]
\centering
\begin{tabular}{|l|l|l|}
\hline
Experiment & BEDS Start Date & BEDS Units \\ \hline
1 & 2020-07-27 & PICUs and PCUs \\ \hline
2 & 2019-01-01 & PICUs and PCUs \\ \hline
\end{tabular}
\caption{Experiment setups in the simulation model for demonstrating the outcome of the implementation of BEDS. In each experiment, elective surgical patients to be admitted to one of the location in \textit{'BEDS Units'} are scheduled by the BEDS algorithm after the \textit{'BEDS Start Date'}.}
\label{table: experiments}
\end{table}
We compare the daily elective surgical admissions into PICUs and PCUs in historical and simulated data in Figure \ref{fig:PICU21_0} and Figure \ref{fig:PICU21}. As a result, we conclude that the simulation model is capable of reproducing historical data in the hospital record during 2019.1.1 to 2020.7.26 before the BEDS implementation. And during the period of 2020.7.27 to 2021.3.31, after the BEDS implementation, there are discrepancies between the actual data in the hospital record and simulated data. The simulation environment may be optimistic when estimating the real-world performance of BEDS. In real-world deployment, the actual performance could be less significant than the simulated one due to implementation details like human execution errors of hospital schedulers executing the scheduling logic, deviations from the assumptions on patient available windows and surgeon availability (as explained in detail in Section \ref{assump}), and missing data in the hospital records. The distribution of the difference between the simulated scheduled time and original time in record is in Figure \ref{fig: diff}.

 In the second experiment, we simulate the potential outcome of implementing BEDS at LPCH at the beginning of 2019. The result was presented to the administrative team of LPCH and they decided to adopt BEDS as their scheduling policy after reviewing the simulated outcome. According to the simulation results summarized in Table \ref{table: PICU19}, the implementation of BEDS in 2019 could have reduced the coefficient of variation (the standard deviation divided by the mean) of the daily elective surgical admission by 35.2$\%$ in PICUs and by $26.0\%$ in PCUs in that year. The hospital had a surgical level loading goal of decreasing the percentage of the days with less than two or more than five daily elective surgical admissions in PICUs. According to the simulation results, if BEDS had been implemented in 2019, the total number of days with less than two or more than five daily elective surgical admissions in PICUs would have been reduced from 120 to 77 in 2019. 


\begin{figure*}[h]
    \centering
    \includegraphics[width = 17cm]{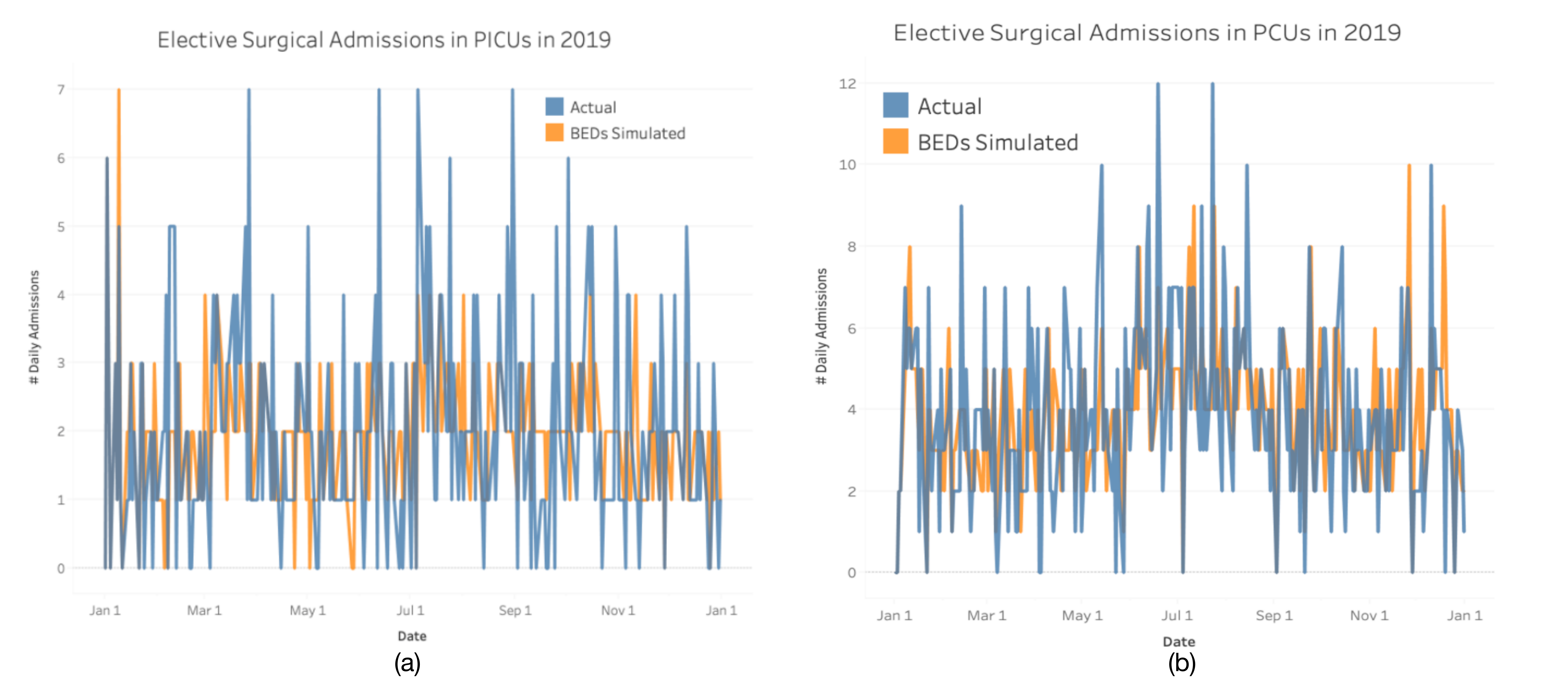}
    \caption{Comparison of the number of daily elective surgical admission in historical data (blue line) and simulated data with BEDS (orange line) in 2019 in PICUs (a) and PCUs (b).}
    \label{figure: PICU19_0}
\end{figure*}


 \begin{figure*}[h]
    \centering
        \includegraphics[width = 15 cm]{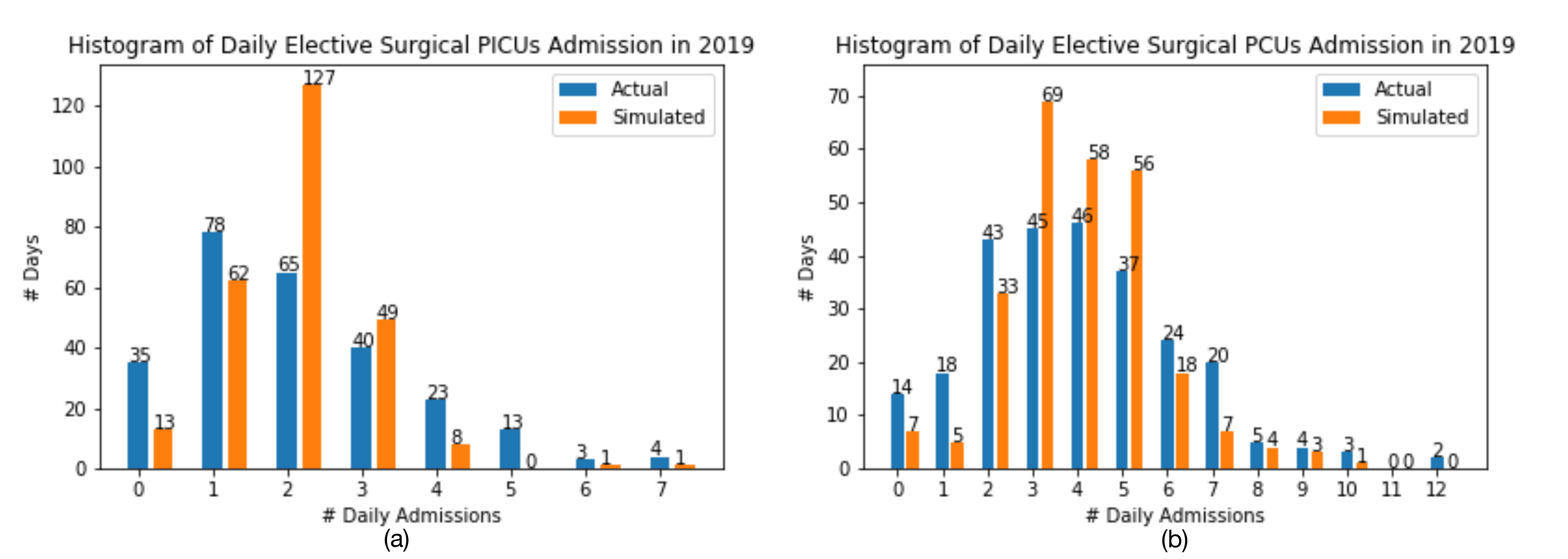}
    \caption{Comparison of the histogram of daily elective surgical admission in historical data (blue bar) and simulated data with BEDS  (orange bar) in 2019 in PICUs (a) and PCUs (b).}
    \label{figure: PICU19}
\end{figure*} 

 \begin{table}[h]
\centering
\begin{tabular}{|l|l|l|}
\hline
PICUs & Historical & BEDS Simulated \\ \hline
Mean & 2.034 & 1.946 \\ \hline
CoV & 0.754 & 0.488 \\ \hline
$90\%$ Quantile/Median & 2.0 & 1.5 \\ \hline
PCUs & Historical & BEDS Simulated \\ \hline
Mean & 3.916 & 3.893 \\ \hline
CoV & 0.577 & 0.427 \\ \hline
$90\%$ Quantile/Median& 1.75 & 1.5 \\ \hline
\end{tabular}
\caption{Comparison of the historical data and simulated results with BEDS implemented in 2019 for daily elective surgical admissions in PICUs and PCUs.}
\label{table: PICU19}
\end{table}

\section{Implementation in Hospital}
\label{sec:implementation}
The BEDS algorithm was incorporated into the scheduling system of the Stanford children's hospital on 2020-07-27 for all elective surgical patients that were known, at the time of scheduling, to be cases that would require further admission to the pediatric intensive care units (PICUs) or acute care units (PCUs).

The implementation of BEDS required buy-in from various stakeholders throughout the health system. A team of surgeons, schedulers, information services representatives, and peri-operative leadership finalized a design for the tool that did not require a significant change to the existing surgery scheduling process. BEDS was built over a six-month period directly in EPIC, Stanford's electronic medical record system. 

When a patient is identified by the surgeon as needing surgery in clinic, the surgeon will contact their scheduler with details about the procedure, urgency of the case, and expected postoperative recovery unit. The scheduler will open the surgical scheduling flowsheet in EPIC and begin to fill in the necessary information from the surgeon. Schedulers will input the postoperative unit and a possible future date for surgery in the scheduling flowsheet. This will automatically populate a calendar heatmap in EPIC that will show dates two weeks prior to the reference date and a month post the reference date. Using the heatmap as a guide, schedulers will select a few days that have a low number of surgery admissions to propose to the patient. The schedulers will then contact the patient to confirm a date of surgery.  

The main function of BEDS is the heatmap to visualize the best and worst days to schedule patients for surgery based on the number of currently scheduled surgical patients with the same in-hospital postoperative recovery destination. The BEDS heatmap also works for scheduling outpatient surgeries, cases that do not require a hospital bed post-operation. BEDS is used by all surgical services, with the exception of Cardiology, to schedule elective surgical cases. Design of the functionality and features of BEDS required input from stakeholders throughout the patient journey, especially the end users (i.e. the surgery schedulers). Challenges with leveling the number of hospital beds needed for surgery patients were handled by the operating room teams on the day of surgery. With BEDS giving visibility to future hospital bed needs, level loading challenges are now solved at the time of scheduling the surgical case. BEDS has been designed so the ranges for the colors within the heatmap can be changed if adjustments need to be made as surgical volumes grow. 

\begin{figure*}[h]
    \centering
\includegraphics[width = 13 cm]{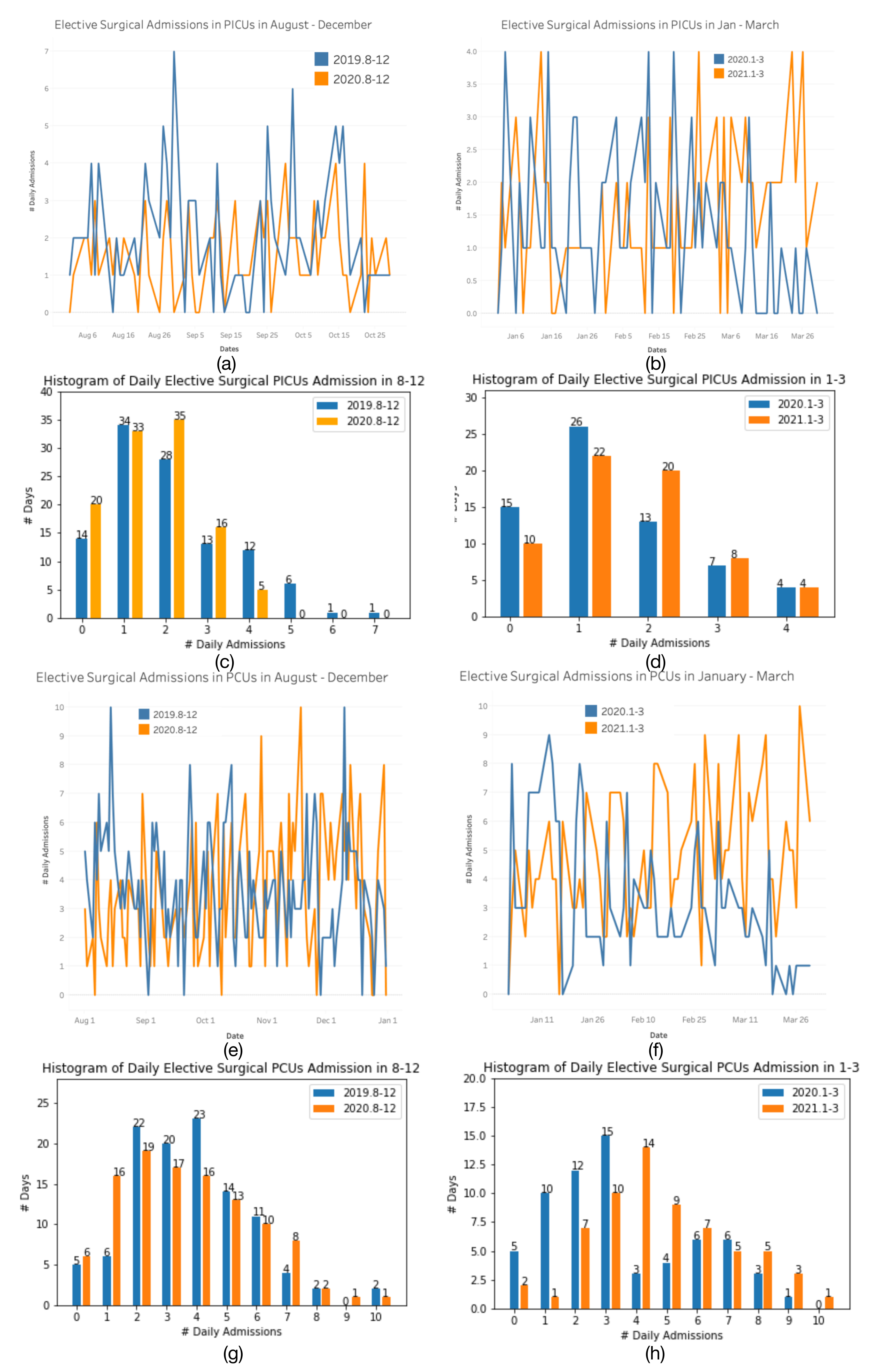}
\caption{Comparison between periods during the same time of year before and after the implementation of BEDS. We show the number of daily elective surgical admission in 2019.8-12 (blue line) and 2020.8-12 (orange line) in PICUs in (a). We also compare their histograms in (c). At the same time,  we compare the number of daily elective surgical admission in 2020.1-3 (blue line) and 2021.1-3 (orange line) in PICUs in (b) and compare their histograms in (d). Similarly, for PCUs, we show the number of daily elective surgical admission in 2019.8-12 (blue line) and 2020.8-12 (orange line) in PCUs in (e) and compare their histograms in (g). Meanwhile, we compare the number of daily elective surgical admission in 2020.1-3 (blue line) and 2021.1-3 (orange line) in PCUs in (f) and compare their histograms in (h).}
    \label{fig: BEDs2020}
\end{figure*}

To demonstrate the impact of implementing BEDS at the Stanford Children's Hospital, we present a comparison between the period of 2020-8 to 2021-3 after BEDS and the same months in the previous year (2019-8 to 2020-3) before BEDS in Table \ref{table: BEDs2020}, \ref{table: BEDs2020_PCU} and Figure \ref{fig: BEDs2020}. We compare the same months during the year to accommodate seasonality issues. We compare the histogram, the mean, the coefficient of variation, the median, the $90\%$ quantile, and the $90\%$  quantile - median ratio of the number of daily elective surgical admissions in PICUs and PCUs before and after BEDS. 

\begin{table*}[h]
\centering
\begin{tabular}{|l|l|l|l|l|}
\hline
Periods & 2019.8-12  & 2020.8-12 & 2020.1-3 & 2021.1-3 \\ \hline
BEDS & before & after & before & after \\ \hline
Mean & 2.018 & 1.569 & 1.375 & 1.594 \\ \hline
CoV & 0.744 & 0.693 & 0.828 & 0.681 \\ \hline
Median & 2.0 & 2.0 & 1.0 & 1.5 \\ \hline
$90\%$ Quantile & 4.0 & 3.0 & 3.0 & 3.0 \\ \hline
$90\%$ Quantile - Median Ratio & 2.0 & 1.5 & 3.0 & 2.0 \\ \hline
\end{tabular}
\caption{Comparison of the number of daily elective surgical admissions to PICUs between periods during the same time of year before and after the implementation of BEDS.}
\label{table: BEDs2020}
\end{table*}

\begin{table*}[h]
\centering
\begin{tabular}{|l|l|l|l|l|}
\hline
Periods & 2019.8-12  & 2020.8-12 & 2020.1-3 & 2021.1-3 \\ \hline
BEDS & before & after & before & after \\ \hline
Mean & 3.688 & 3.532 & 3.415 & 4.688 \\ \hline
CoV & 0.538 & 0.623 & 0.695 & 0.478 \\ \hline
Median & 4.0 & 3.0 & 3.0 & 4.0 \\ \hline
$90\%$ Quantile & 6.0 & 7.0 & 7.0 & 8.0 \\ \hline
$90\%$ Quantile - Median Ratio & 1.500 & 2.333 & 2.333 & 2.000 \\ \hline
\end{tabular}
\caption{Comparison of the number of daily elective surgical admissions to PCUs between periods during the same time of year before and after the implementation of BEDS.}
\label{table: BEDs2020_PCU}
\end{table*}

\begin{figure*}[h]
    \centering
    \includegraphics[width = 7 cm]{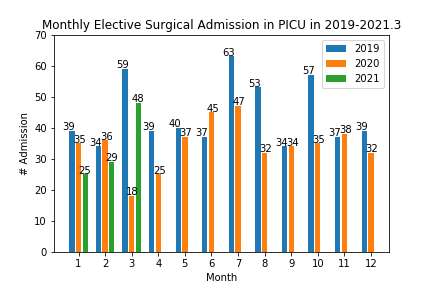}
    \caption{Monthly elective surgical admissions in PICUs in 2019.1-2020.3.}
    \label{fig: Vol}
\end{figure*}

We notice that the volume of the patient flow at the Stanford Children's Hospital has been fluctuating in 2020 due to the outbreak of COVID-19 and other changes at the hospital, as reflected in Figure \ref{fig: Vol}. Hence, to compare different periods before and after BEDS, we focus on a metric that is not proportional to the volume of the patient flow:  the $90 \%$ quantile to median ratio of the daily elective surgical admission (QMRA). As in Table \ref{table: BEDs2020}, the QMRA in PICUs dropped from 2.0 (2019.8-12) to 1.5 (2020.8-12) and from 3.0 (2020.1-3) to 2.0 (2021.1-3) after BEDS implementation. In PCUs, the QMRA changed from 1.500 (2019.8-12) to 2.333 (2020.8-12) and dropped from 2.333 (2020.1-3) to 2.000 (2021.1-3) after BEDS implementation. As the result shows, it took a warm-up period in PCUs for BEDS to start to take effect.

The daily admissions of elective surgical patients on different weekdays follow different patterns due to the arrangement of the block schedule of the surgeons, as described in Section \ref{sec:intro}. Hence, we partition the weekdays into five groups and compare the QMRA in each group before (2019.8 - 2020.3) and after (2020.8 - 2021.3) BEDS as in Table \ref{table: hp_PICU} and Table \ref{table: hp_PCU}. Within the two periods that we are comparing, there are in total $n=34$ weeks. As a result, in PICUs, the QMRA dropped on Mondays, Tuesdays, and Thursdays, which are the three weekdays with the highest ratios before BEDS. And no weekday has a QMRA higher than 2.0 after BEDS. Similarly, in PCUs, the QMRA dropped on Mondays, Tuesdays and Wednesdays, which are the three weekdays with the highest ratios before BEDS. We propose a one-sided hypothesis test for the size of the change of the QMRA after BEDS. For  the  distribution of the test statistics, we estimate it by non-parametric bootstrapping with $m= 100,000$ samples. The results are summarized in Table \ref{table: hp_PICU} and Table \ref{table: hp_PCU}. For PICUs, the null hypothesis is that the change of the QMRA in the post-BEDS period (2020.8.1 - 2021.3.31) compared with the same time during the  previous  year  before  BEDS (2019.8.1 - 2020.3.31) is less than 0.25. For PICUs, the p-value of the hypothesis test is $0.0778$ on Thursdays and $0.0807$ on Mondays. This means that there is significant evidence that we can safely reject the null hypothesis . Therefore there is evidence that there is a significant improvement of the QMRA corresponding to a drop more than $25\%$ on Thursdays and Mondays after BEDS. For the other weekdays, there is no evidence of a significant change of the QMRA. Overall, noticing that Thursdays and Mondays are the two weekdays with the highest QMRA before BEDS, BEDS improved the patient flow on the two most conjested weekdays in PICUs. For PCUs, the null hypothesis is that the change of the QMRA in the post-BEDS period (2020.8.1 - 2021.3.31) compared with the same time during the  previous  year  before  BEDS (2019.8.1 - 2020.3.31) is less than 0.1. For PCUs, the p-value of the hypothesis test is $0.342$ on Tuesdays and $0.410$ on Mondays. Hence there is no evidence of a significant change of the QMRA after BEDS in PCUs. In conclusion, based on these results, overall, BEDS improves the performance on these metrics.

\begin{table*}
\centering
\begin{tabular}{|l|l|l|l|l|l|}
\hline
90\% Quantile to Median Ratio/Weekday & Monday  & Tuesday & Wednesday & Thursday & Friday \\ \hline
2019.8 - 2020.3 & 3.000 & 2.700 & 1.667 & 3.700 & 1.500 \\ \hline
2020.8 - 2021.3 & 2.000 & 2.000 & 2.000 & 1.500 & 2.000 \\ \hline
Direction of Change & - & - & + & - & + \\ \hline
p-value (Size of Change $\le 0.25$) & 0.0807 & 0.385 & 0.504 & 0.0778 & 0.415 \\ \hline
\end{tabular}
\caption{Comparison of the $90\%$ quantile - mean ratio of the number of daily elective surgical admissions to PICUs between periods during the same time of year before and after the implementation of BEDS. }
\label{table: hp_PICU}
\end{table*}

\begin{table*}
\centering
\begin{tabular}{|l|l|l|l|l|l|}
\hline
90\% Quantile to Median Ratio/Weekday & Monday  & Tuesday & Wednesday & Thursday & Friday \\ \hline
2019.8 - 2020.3 & 2.233 & 2.000 & 1.914 & 1.750 & 1.429 \\ \hline
2020.8 - 2021.3 & 2.000 & 1.400 & 1.711 & 1.925 & 1.667 \\ \hline
Direction of Change & - & - & - & + & + \\ \hline
p-value (Size of Change $\le 0.1$) & 0.410 & 0.342 & 0.547 & 0.710 & 0.708 \\ \hline
\end{tabular}
\caption{Comparison of the $90\%$ quantile - mean ratio of the number of daily elective surgical admissions to PCUs between periods during the same time of year before and after the implementation of BEDS. }
\label{table: hp_PCU}
\end{table*}

\section{BEDS Framework} \label{Section_BEDS_Frame}

The BEDS algorithm can be seen as one particular instance of a general BEDS framework \ref{algorithm:BEDS} designed to produce algorithms with several desirable properties. The most important such property is that any algorithm generated by the BEDS framework is compatible with a very common surgical scheduling workflow: a surgeon examines a patient in clinic, recommends a type of surgery, and estimates how soon it should be performed. The patient then works with a scheduler to choose a day of surgery. Based on the constraints and objectives of the institution, the BEDS framework can be used to produce an algorithm to help the scheduler choose which dates to offer the patient. The only data required are the same as those for the BEDS algorithm (clinically acceptable wait time for the procedure, surgeon availability, patient availability, estimated procedure duration, and occupancy data for the post-operative unit), but additional data may be used for more sophisticated scheduling. We provide pseudo-code to explain the BEDS framework.

The ranking algorithm chosen for the modifiable step may be fully-interpretable or based on a sophisticated optimization model. In either case, the results may be used to generate a simple, color coded calendar such as Figure \ref{fig: BEDs}. The amount of choice patients have in scheduling their procedure may be altered by modifying the parameter $n$ and offering each patient a choice amongst the top $n$-ranked days. 

\section{Discussion and Conclusions}
\label{sec:discussion}

We developed and deployed BEDS (better elective day of surgery) a flexible, interpretable algorithm to assist in the selection of the day for which a surgical procedure is to be scheduled. When deployed at an academic medical center, the use of BEDS over an 8 month period reduced variation in the number of daily elective surgical admissions compared to the same 8 months in earlier years. We showed that BEDS relies on very few, commonly collected data points, offers fully interpretable recommendations, and is compatible with implementation in the very technically limited functionality commonly available in large EMRs. We then introduced the BEDS framework, a general algorithmic framework of which BEDS is one application. The algorithms generated by the BEDS framework may retain many of the desirable characteristics of BEDS while being compatible with a wide range of objectives and constraints. 

The evaluation of the deployment of BEDS should be interpreted in light of a significant limitation. The disruption caused by COVID-19 changed healthcare provision, especially elective surgical care, in numerous ways for which we could not control.

The BEDS algorithm is ready and available for use by other institutions and freely available for download as a Tableau workbook. The BEDS framework may be useful for institutions seeking to design a readily implementable approach to improving scheduling or as a performance benchmark for researchers developing more sophisticated scheduling algorithms. 

\subsection{Acknowledgments}
We thank numerous members of the hospital leadership for their partnership in implementing BEDS. In particular, Dr. Rebecca Claure, Dr. James Dunn, Patrick Kane, Ryan Bruvold, and Derek Garnholz. 


\section*{Declarations}
The authors report no conflicts or external funding.

\begin{algorithm}[H]
\SetAlgoLined
    For each day $d$ in the scheduling horizon $D$, require: 
    \begin{enumerate}
        \item For each surgeon $j$, their number of available hours $h_{d,j}$.
        \item For each post-op unit $u$, the number of patients $n_{d,u}$ scheduled to be admitted.
        \item \textbf{Optional} Additional characteristics of the institution $h_d$.
    \end{enumerate}
    
\For{patient $i$ being scheduled for a procedure}{
    Determine that:\\
    \text{\quad}1. Maximum acceptable wait time for the 
    \text{\quad}procedure.\\ 
    \text{\quad}2. Surgeon $j$ that will perform the proce-\\
    \text{\quad}dure.\\
    \text{\quad}3. Scheduled procedure duration, $g_{i}$.\\
    \text{\quad}4. Patient's availability window for \\
    \text{\quad}procedure, $[l_i, r_i]$.\\
    \text{\quad}5. Unit $u_i$ to which the patient will be \\
    \text{\quad}admitted after the procedure.\\
    \text{\quad}6. \textbf{Optional} Additional patient characteris-
    \\\text{\quad}tics $p_i$.\\
    Based on the above:\\
        \text{\quad}1. The candidate dates $\mathcal{D}_i$ are those within\\ \text{\quad}the maximum acceptable wait time on which \\
        \text{\quad}the surgeon has sufficient time, i.e., $h_{d,j} \ge\\ \text{\quad}g_{i}$.\\
        \text{\quad}2. (\textbf{Modifiable ranking}) Rank the days in\\ \text{\quad}$d\in\mathcal{D}_i$ based on an objective function and\\ \text{\quad}constraints based on $j$, $h_{d,j}$, $g_i$, $p_i$, and\\ \text{\quad}$[l_i, r_i]$.\\
        \text{\quad}3. Return the $n$ top-ranked candidate dates.\\
        \text{\quad}4. Update $h_{d_{i},j}$, $n_{d_{i},u}$, and $h_d$.
}
\caption{Algorithmic Framework: BEDS}
\label{algorithm:BEDS}
\end{algorithm}










\begin{appendices}
\renewcommand\thefigure{\arabic{figure}}
\setcounter{figure}{10}
\renewcommand\thetable{\arabic{table}}
\setcounter{table}{13}
\section{}
\label{sec:appendix}

\begin{figure*}[h]
    \centering
    \includegraphics[width = 13 cm]{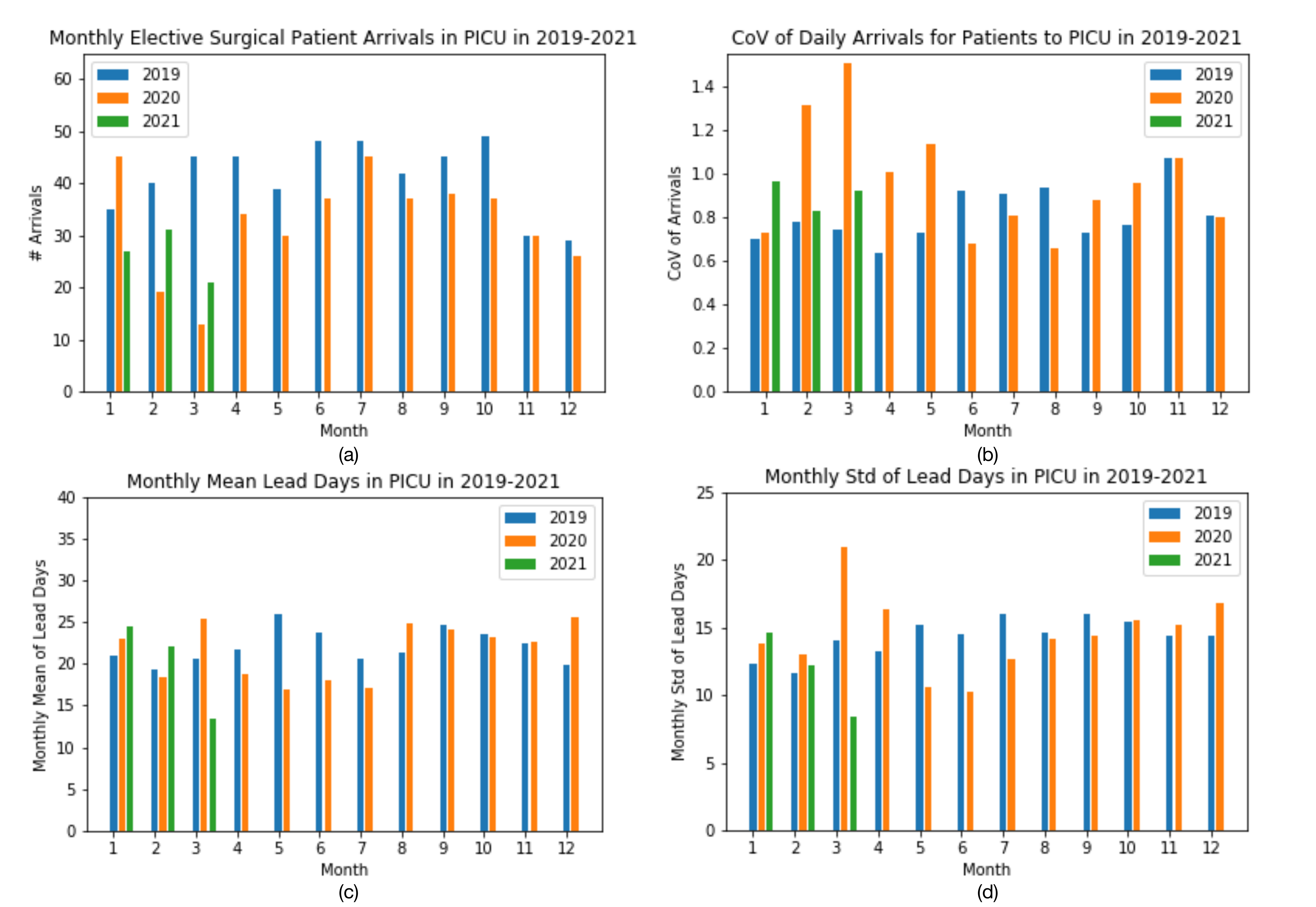}
    \caption{Arrivals and lead days of elective surgical PICUs patient in 2019-2021.3. For the period of 2019-2021.3, we show the monthly mean of the daily arrivals of elective surgical PICUs patients (a), the coefficient of variation of the daily arrivals of elective surgical PICUs patients in every month (b), the monthly mean of the number of lead days of elective surgical PICUs patients (c), and the standard deviation of the lead days of elective surgical PICUs patients in every month (d).}
    \label{fig: back}
\end{figure*}

\begin{figure*}[h]
    \centering
    \includegraphics[width = 7 cm]{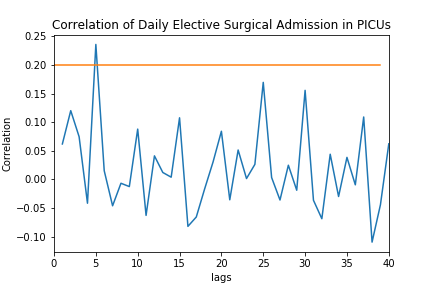}
    \caption{Correlation of the daily elective surgical admissions in PICUs with neighboring days.}
    \label{fig: cor}
\end{figure*}

As a background of the variability of the volume of the demand at the hospital during different periods we are comparing in Section \ref{sec:implementation}, we present statistics of the daily elective surgical patient arrivals in PICUs as well as the lead time in each month during 2019.1 - 2021.3 in Figure \ref{fig: back}. As in Figure \ref{fig: cor}, we calculate the auto-correlation of the daily elective surgical admissions in PICUs on weekdays and find that the auto-correlation peaks regularly for the lags of 5, 10 and 15 days, etc. This phenomenon can be explained by the weekly surgeon OR time block described in Section \ref{sec:intro}. 


\subsection{More on the simulation model}
\label{sec: simulation model more}
In this section, we will explain more details of the simulation model. 
\subsubsection{Data inputs and pre-processing}
\label{app_sim}
Below, we show the headers of the data \textit{midnight census}, \textit{procedure record}, and \textit{surgeon availability} in Table \ref{table:census-header}, \ref{table:proc-header} and \ref{table:surgeon-avail-header} as examples of the inputs. In addition, we point out in the column \textit{'Required'} if each column in the headers is a required input for using the discrete event simulation model. The unrequired columns can be useful for either model validation or follow-up analysis of the results and are recommended during data collection. The column \textit{'Primary CSN'} is an encoded patient-visit ID. Each patient has a unique ID for each time he/she visits the hospital. 

\begin{table*}[h]
\centering
\begin{tabular}{|l|l|l|}
\hline
Required & Column & Value \\ \hline
Yes & Primary CSN & 1  \\ \hline
Yes & Dept Abbrev & PICUs \\ \hline
Yes & Effective Date/Time & 15-1-1 23:59:00 \\ \hline
Yes & Hospital Admission Dt/Tm & 15-1-1 0:10:00 \\ \hline
No & Hospital Discharge Dt/Tm & 15-1-2 14:10:00 \\ \hline
No & Service & General Pediatrics \\ \hline
No& Admit Type & Elective \\ \hline
No & Admit Source & Home \\ \hline
\end{tabular}
\caption{Midnight census data header. The column \textit{`Dept Abbrev'} is the location of the patient in record at midnight. The date of one census record can be read from the column \textit{`Effective Date/Time'}. }
\label{table:census-header}
\end{table*}

\begin{table*}[h]
\centering
\begin{tabular}{|l|l|l|}
\hline
Required & Column & Value \\ \hline
Yes & Primary CSN & 2 \\ \hline
Yes &Primary Surgeon ID & 3\\ \hline
Yes &Location & MAIN OR \\ \hline
Yes & Originally Scheduled On & 57-11-2 16:02:00 \\ \hline
Yes & Originally Scheduled For & 58-1-1 14:05:00 \\ \hline
Yes &Patient in Room & 58-1-1 13:00:00 \\ \hline
Yes & Patient out of Room & 58-1-1 15:02:00 \\ \hline
No &Patient Class & Outpatient Surgery \\ \hline
No &Service & Otolaryngology \\ \hline
No &Primary Procedure ID & 4 \\ \hline
\end{tabular}
\caption{Procedure record data header. The column \textit{'Originally Scheduled On'} is the time the patient comes to the hospital to schedule his/her visit. The column \textit{'Originally Scheduled For'} is the time the patient's surgery is scheduled for. The two columns are used to calculate the patient's lead days. The column \textit{'Patient in Room'} records the time surgeries start and \textit{'Patient out of Room'} shows when surgeries end. The column \textit{'Patient Class'} is a reference for inferring if the patient is a elective patient or not. It is useful for identifying the surgical outpatients. More details in data-preprocessing.} 
\label{table:proc-header}
\end{table*}

\begin{table}[h]
\centering
\begin{tabular}{|l|l|l|}
\hline
Required & Column & Value \\ \hline
Yes & Date & 2001-01-01 \\ \hline
Yes & Primary Surgeon ID & 20 \\ \hline
Yes & Service & Cardiology \\ \hline
Yes & Available Hours & 3.2\\ \hline
\end{tabular}
\caption{Surgeon availability data header. The column '\textit{Available Hours'} is the number of hours a surgeon is available for performing surgeries on a given day.}
\label{table:surgeon-avail-header}
\end{table}

With the data inputs in Table \ref{table: inputs} provided, the simulation system will first pre-process them with the following steps:

\begin{itemize}
    \item Scope of Units: focus on Pediatric Areas (Acute and ICU areas).
    \item Warm-up Period: We include a warm-up period of one year. For example, if the goal is to simulate the patient flow starting from 2018-01-01, we include all patients in record who scheduled their visit or arrived without appointment no earlier than 2017-01-01 to warm up the model. 
    \item Data Cleaning: we exclude the following types of patients and records:
    \begin{itemize}
        \item patients with duplicate census record on the same day or inconsecutive census records during their stay,
        \item patients with overnight surgeries,
        \item surgery records missing the column \textit{'Patient in Room'} or \textit{'Patient out of room'}.
    \end{itemize}
\end{itemize}

\begin{table*}[h]
\centering
\begin{tabular}{|l|l|}
\hline
Primary CSN & 3 \\ \hline
Patient Class & Surgery Admit \\ \hline
Arrival Time & 3018-02-22 12:51 \\ \hline
Admission Time & 3018-04-01 11:20 \\ \hline
Available Window & [3018-02-23 12:51, 3018-05-27 11:59] \\ \hline
Unit List & [MAIN OR, PCUs] \\ \hline
LOS List & [0 days 01:50, 4 days 00:00] \\ \hline
In OR Times & [3018-04-10 11:20] \\ \hline
Primary Surgeon ID & 4 \\ \hline
\end{tabular}
\caption{Patient profile header. The column \textit{`Patient Class'} can take values as 'Surgical Outpatient', 'Surgical Admit', 'Surgical Inpatient', or 'Medical Inpatient'. A patient classified to be 'Surgical Outpatient' or 'Surgical Admit' is an elective surgical patient. The column \textit{`Arrival Time'} is the time a patient comes to schedule his/her visit or arrives at the hospital without an appointment. The column \textit{`Admission Time'} is the time of a patient's admission into the hospital. The column \textit{`Available Window'} is the inferred period that a patient is available for surgery. The column \textit{`Unit List'} is a list of the units that a patient visits during his/her stay in a timely order. The column \textit{`LOS List'} is his/her length of stay in each unit. The column \textit{`In OR Times'} is a list of timestamps that a patient receives surgeries. The column \textit{`Primary Surgeon ID'} records the ID of the primary surgeon of a patient's first surgery at the hospital.} 
\label{table:patient-profile-header}
\end{table*}

We re-organize the input data into two data sets: \textit{patients} and \textit{surgeon availability}. The \textit{patient profile} data contains the information on each patient visit to the hospital, as shown in Table~\ref{table:patient-profile-header}. The \textit{surgeon availability} data record the available times of the surgeons, as in Table \ref{table:surgeon-avail-header}. When \textit{surgeon availability} is not provided, the model has a built-in function that infers \textit{surgeon availability} with the following logic: The surgeon is assumed to be available at a specific date with a full block size, for example, seven hours, if the total number of hours of scheduled cases on record on the same day exceeds a threshold, for example, three hours. The threshold and full block size are inputs that users can adjust by themselves.

\subsubsection{Assumptions}
\label{assump}
We present here a list of assumptions we make relative to the patient flow and the hospital policy when designing the simulation model.
\begin{itemize}
    \item (Elective patients) A patient is an elective surgical patient and can be rescheduled if he/she satisfies one of the following criteria:
    \begin{enumerate}
        \item The patient is a surgical outpatient or surgical admit, 
        \item The patient's lead time is no less than one day.
    \end{enumerate}
    \item (Available window) For an elective surgical patient whose arrival date is $d_1$ and his/her admission date is $d_2$ $(d_2>d_1)$, the length of the patient's lead day is $d_{l}=d_2-d_1$. If the date the patient originally receives his/her surgery is $d_{0}$, we assume that the available window for the patient to receive surgery is $[\max(d_{1}+1, d_0-\alpha*d_{l}),  d_0+\alpha*d_{l}]$. Here, $\alpha$ is a scaling factor of the patient's availability window and is one by default.
    \item (Reschedule by first surgery) The timestamp that a patient is scheduled to by BEDS only depends on his/her first surgery.
    \item Rescheduling a patients dose not change the list of units the patient visits or the length of stay in each unit.
\end{itemize}

\subsubsection{Model Structure}
We introduce more details on our discrete event simulation (DES) model. Figure 2 shows the outline of the DES model. The model consists of three major components:

\begin{enumerate}
    \item Pre-processor: the processor prepares and re-organizes the input \textit{midnight census} data and \textit{procedure record} data into \textit{patients} and \textit{surgeon availability} as described in \ref{app_sim}. 
    
    \item Scheduler (A module encoding the scheduling logic): It is called every time a scheduling request is made, and generates a specific scheduled time following a given scheduling rule. Currently, the scheduler has two modes: historical data and BEDS. The logic of BEDS is encoded into the scheduler, so that the simulator is able to generate the potential outcome associated with implementing BEDS.
    
    \item Event queue: It is a priority queue with time being the index. In other words, the head of the queue will always be the event that has the earliest time. The idea of DES is implemented utilizing this queue, where the simulator repetitively reads from the head of the event queue, pops this event from the queue, generates the next event following a given logic, and pushes the new event into the queue. The event types we consider are summarized in Table~\ref{table:events}. Each event is associated with a specific patient.
\end{enumerate}

\begin{figure*}[h]
    \centering
    \includegraphics[width = 15 cm]{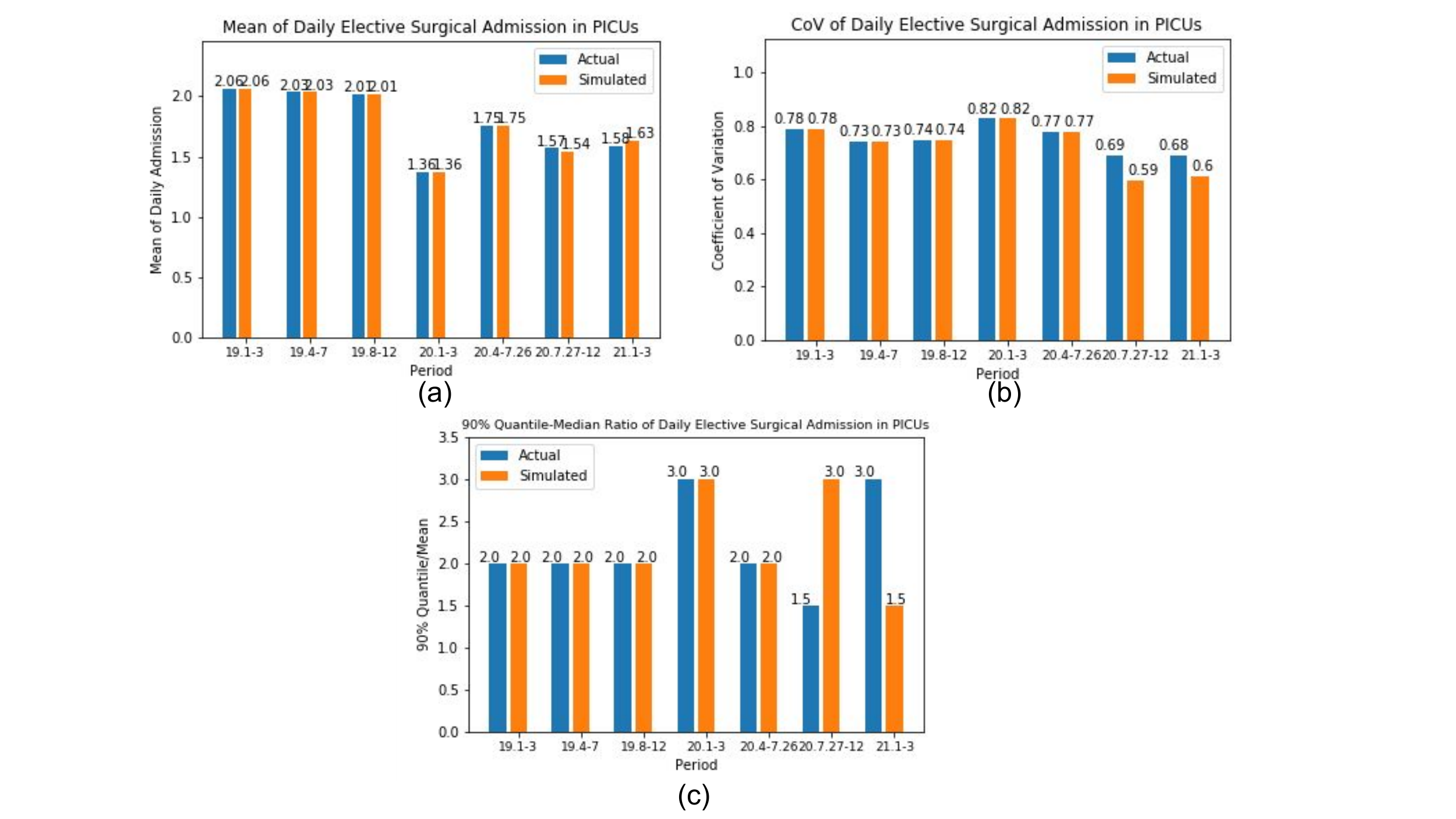}
    \caption{In this figure, we split the period of 2019.1 - 2021.3 into seven time intervals: 19.1-3, 19.4-7, 19.8-12, 20.1-3, 20.4-7.26 (before BEDS), and 20.7.27-12, 21.1-3 (after BEDS). We show the comparison of the mean (a), the coefficient of variation (b) and the 90\% quantile - median ratio (c) of the number of daily elective surgical patient admissions into PICUs in actual (blue bars) and simulated data (orange bars) during the seven periods mentioned above.}
    \label{fig:PICU21_1}
\end{figure*}

We present more results for the simulation model validation in Section \ref{sec:results}. For the first experiment in Section \ref{sec:results}, the comparison of the mean, coefficient of variation, and the $90\%$ quantile - median ratio of daily elective surgical admission in PICUs between historical data and simulated data are in Figure \ref{fig:PICU21_1}. 


\end{appendices}


\bibliography{sn-bibliography}

\begin{thebibliography}{19}
\providecommand{\natexlab}[1]{#1}
\providecommand{\url}[1]{{#1}}
\providecommand{\urlprefix}{URL }
\providecommand{\doi}[1]{\url{https://doi.org/#1}}
\providecommand{\eprint}[2][]{\url{#2}}
 \bibcommenthead

\bibitem[{Bekker and Koeleman(2011)}]{bekker2011scheduling}
Bekker R, Koeleman PM (2011) Scheduling admissions and reducing variability in
  bed demand. Healthcare Management Science 14(3):237

\bibitem[{Beli{\"e}n and Demeulemeester(2007)}]{belien2007building}
Beli{\"e}n J, Demeulemeester E (2007) Building cyclic master surgery schedules
  with leveled resulting bed occupancy. European Journal of Operational
  Research 176(2):1185--1204

\bibitem[{Beli{\"e}n et~al(2009)Beli{\"e}n, Demeulemeester, and
  Cardoen}]{belien2009decision}
Beli{\"e}n J, Demeulemeester E, Cardoen B (2009) A decision support system for
  cyclic master surgery scheduling with multiple objectives. Journal of
  Scheduling 12(2):147--161

\bibitem[{van~den Broek~d’Obrenan et~al(2020)van~den Broek~d’Obrenan,
  Ridder, Roubos, and Stougie}]{van2020minimizing}
van~den Broek~d’Obrenan A, Ridder A, Roubos D, et~al (2020) Minimizing bed
  occupancy variance by scheduling patients under uncertainty. European Journal
  of Operational Research 286(1):336--349

\bibitem[{Childers and Maggard-Gibbons(2018)}]{childers2018understanding}
Childers CP, Maggard-Gibbons M (2018) Understanding costs of care in the
  operating room. JAMA surgery 153(4):e176,233--e176,233

\bibitem[{Chow et~al(2011)Chow, Puterman, Salehirad, Huang, and
  Atkins}]{chow2011reducing}
Chow VS, Puterman ML, Salehirad N, et~al (2011) Reducing surgical ward
  congestion through improved surgical scheduling and uncapacitated simulation.
  Production and Operations Management 20(3):418--430

\bibitem[{Fairley et~al(2019)Fairley, Scheinker, and
  Brandeau}]{fairley2019improving}
Fairley M, Scheinker D, Brandeau ML (2019) Improving the efficiency of the
  operating room environment with an optimization and machine learning model.
  Healthcare Management Science 22(4):756--767

\bibitem[{Guerriero and Guido(2011)}]{guerriero2011operational}
Guerriero F, Guido R (2011) Operational research in the management of the
  operating theatre: a survey. Healthcare Management Science 14(1):89--114

\bibitem[{Jebali and Diabat(2017)}]{jebali2017chance}
Jebali A, Diabat A (2017) A chance-constrained operating room planning with
  elective and emergency cases under downstream capacity constraints. Computers
  \& Industrial Engineering 114:329--344

\bibitem[{Neyshabouri and Berg.(2017)}]{two-stage}
Neyshabouri S, Berg. BP (2017) Two-stage robust optimization approach to
  elective surgery and downstream capacity planning. European Journal of
  Operational Research 260(1):21--40

\bibitem[{Papanicolas et~al(2018)Papanicolas, Woskie, and
  Jha}]{10.1001/jama.2018.1150}
Papanicolas I, Woskie LR, Jha AK (2018) {Health Care Spending in the United
  States and Other High-Income Countries}. JAMA 319(10):1024--1039.
  \doi{10.1001/jama.2018.1150},
  \urlprefix\url{https://doi.org/10.1001/jama.2018.1150},
  {\href{https://arxiv.org/abs/https://jamanetwork.com/journals/jama/articlepdf/2674671/jama\_papanicolas\_2018\_sc\_180001.pdf}{{https://arxiv.org/abs/https://jamanetwork.com/journals/jama/articlepdf/2674671/jama\_papanicolas\_2018\_sc\_180001.pdf}}}

\bibitem[{Price et~al(2011)Price, Golden, Harrington, Konewko, Wasil, and
  Herring}]{price2011reducing}
Price C, Golden B, Harrington M, et~al (2011) Reducing boarding in a
  post-anesthesia care unit. Production and Operations Management
  20(3):431--441

\bibitem[{Rahimi and Gandomi(2020)}]{rahimi2020comprehensive}
Rahimi I, Gandomi AH (2020) A comprehensive review and analysis of operating
  room and surgery scheduling. Archives of Computational Methods in Engineering
  pp 1--22

\bibitem[{Samudra et~al(2016)Samudra, Van~Riet, Demeulemeester, Cardoen,
  Vansteenkiste, and Rademakers}]{samudra2016scheduling}
Samudra M, Van~Riet C, Demeulemeester E, et~al (2016) Scheduling operating
  rooms: achievements, challenges and pitfalls. Journal of scheduling
  19(5):493--525

\bibitem[{Scheinker and Brandeau(2020)}]{scheinker2020implementing}
Scheinker D, Brandeau ML (2020) Implementing analytics projects in a hospital:
  Successes, failures, and opportunities. INFORMS Journal on Applied Analytics
  50(3):176--189

\bibitem[{Shehadeh and Padman(2020)}]{shehadeh2020stochastic}
Shehadeh K, Padman R (2020) Stochastic optimization approaches for elective
  surgery scheduling and downstream capacity planning: Models, challenges, and
  opportunities. SSRN preprint

\bibitem[{Zenteno et~al(2015)Zenteno, Carnes, Levi, Daily, Price, Moss, and
  Dunn}]{zenteno2015pooled}
Zenteno AC, Carnes T, Levi R, et~al (2015) Pooled open blocks shorten wait
  times for nonelective surgical cases. Annals of Surgery 262(1):60--67

\bibitem[{Zenteno et~al(2016)Zenteno, Carnes, Levi, Daily, and
  Dunn}]{zenteno2016systematic}
Zenteno AC, Carnes T, Levi R, et~al (2016) Systematic or block allocation at a
  large academic medical center. Annals of Surgery 264(6):973--981

\bibitem[{Zhu et~al(2019)Zhu, Fan, Yang, Pei, and Pardalos}]{zhu2019operating}
Zhu S, Fan W, Yang S, et~al (2019) Operating room planning and surgical case
  scheduling: a review of literature. Journal of Combinatorial Optimization
  37(3):757--805

\end{thebibliography}
\bibliographystyle{sn-basic}

\end{document}